\documentclass[superscriptaddress, notitlepage]{revtex4-1}

\usepackage{graphicx}
\usepackage{amssymb}
\usepackage{amsthm}
\usepackage{amsmath}
\usepackage{color}
\usepackage[normalem]{ulem}
\usepackage{multirow}
\usepackage{subfigure}
\usepackage{url}

\begin{document}

\title{An Optimized Shape Descriptor Based on Structural Properties of Networks}

\author{Gisele H. B. Miranda}
\email{gimiranda@usp.br}
\affiliation{Institute of Mathematics and Computer Science, University of S\~{a}o Paulo, S\~{a}o Carlos - SP, 13560-970, Brazil.}
\author{Jeaneth Machicao}
\email{machicao@usp.br}
\affiliation{S\~{a}o Carlos Institute of Physics, University of S\~{a}o Paulo, S\~{a}o Carlos - SP, PO Box 369, 13560-970, Brazil.}
\author{Odemir M. Bruno}
\email{bruno@ifsc.usp.br}
\affiliation{S\~{a}o Carlos Institute of Physics, University of S\~{a}o Paulo, S\~{a}o Carlos - SP, PO Box 369, 13560-970, Brazil.}
\affiliation{Institute of Mathematics and Computer Science, University of S\~{a}o Paulo, S\~{a}o Carlos - SP, 13560-970, Brazil.}

\begin{abstract}
The structural analysis of shape boundaries leads to the characterization of objects as well as to the understanding of shape properties. The literature on graphs and networks have contributed to the structural characterization of shapes with different theoretical approaches. We performed a study on the relationship between the shape architecture and the network topology constructed over the shape boundary. For that, we used a method for network modeling proposed in 2009. Firstly, together with curvature analysis, we evaluated the proposed approach for regular polygons. This way, it was possible to investigate how the network measurements vary according to some specific shape properties.  Secondly, we evaluated the performance of the proposed shape descriptor in classification tasks for three datasets, accounting for both real-world and synthetic shapes. We demonstrated that not only degree related measurements are capable of distinguishing classes of objects. Yet, when using measurements that account for distinct properties of the network structure, the construction of the shape descriptor becomes more computationally efficient. Given the fact the network is dynamically constructed, the number of iterations can be reduced. The proposed approach accounts for a more robust set of structural measurements, that improved the discriminant power of the shape descriptors.
\end{abstract}


\maketitle

%\begin{keyword}
%Shape analysis \sep Shape classification \sep Network analysis \sep Pattern recognition


%\end{keyword}

%\end{frontmatter}

%\linenumbers

\section{Introduction}
\label{S:1}

In computer vision, shape boundaries are important attributes that can be used for the characterization and the classification of objects. In shape analysis there are many pattern recognition applications covering different areas, such as neuroscience~\cite{gerig2001shape,torres2004graph,fan2007compare}, agriculture~\cite{aitkenhead2003weed,neto2006plant,
brosnan2002inspection,costa2011shape,hemming2001pa,plotze2005leaf}, medical imaging~\cite{rangayyan2000boundary,heimann2009statistical,selle2002analysis,tsai2003shape}, remote sensing~\cite{benz2004multi,blaschke2010object}, to mention but a few. Over the last decades many methods were proposed in the literature of pattern recognition which are based on classical approaches, such as, Fourier descriptors~\cite{osowski2002fourier,zhang2001comparative}, wavelets~\cite{davatzikos2003hierarchical,soares2006retinal}, fractal dimension~\cite{plotze2005leaf,backes2010shape,backes2012shape,bruno2008fractal}, curvature scale space (CSS)~\cite{mokhtarian2013curvature} among others. These methods support a wide range of applications. In addition, methods based on structural properties of shape boundaries have been drawing attention for classification tasks. Such methods are strongly influenced by graph and network theory~\cite{backes2009complex,backes2010shape,backes2013polygonal,wu2015image,sui2016complex}. Instead of considering the shape boundaries only as a chain of connected points, this new approach also takes advantage of the topological properties of the shape contour.

In a paper published in 2009~\cite{backes2009complex}, Backes \textit{et al.} proposed a method, named as CNDescriptor, for boundary shape analysis based on the connectivity among the contour pixels. In this one-parameter model, each pixel is modeled as a vertex of a graph and the connections between them are established according to that parameter, which represents a distance threshold. For a given threshold value, there is one graph realization and a set of measurements regarding the connectivity of this graph can be obtained. These measurements are the so-called network descriptors which characterize the graph structure, and, consequently, unveil the topological properties of the shape boundary. In the referred paper, the authors also detail some properties of the graphs obtained through the proposed method, e.g., for some thresholds the graphs present a high clustering coefficient, characterizing them as small-world graphs. In addition, the CNDescriptor is invariant to geometric transformations such as rotation and scale and robust for what concerns the presence of noise. This method was also evaluated in the context of image degradation, in which parts of the shape contour were removed. The authors demonstrated that the CNDescriptor is also very robust in such cases, showing promising results. It is also possible to find an extension of the method for texture analysis~\cite{backes2013texture}.

However, in spite of being well suited for shape analysis tasks, the dynamic evolution signature provided by the method CNDescriptor only accounts for degree-related measurements as feature vectors. However, a much more robust set of measurements could improve the classification performance in different shape recognition applications.
%was \textcolor{red}{underestimated} regarding the measurements that can be obtained from the graph structure. The authors identified interesting characteristics of such graphs which are related to the small-world property, but only degree-related measurements were evaluated as feature vectors. 
Along the last two decades, Complex Networks (CN) has been established as a new research field which integrates graph theory and statistical mechanics~\cite{newman2003structure,boccaletti2006complex,newman2011structure}. Many related studies have provided new insights about the topological characterization of networks leading to a deep understanding of how the connectivity patterns of the nodes are related to the network model~\cite{barabasi1999emergence,barabasi2000scale,watts1998collective}. Therefore, the measurements extracted from the network structure are important features for the network characterization. In a survey of 2007, Costa \textit{et al.} emphasize this approach~\cite{costa2007characterization}. The authors summarize a huge set of measurements that can be used to describe the topology of a network. They present different categories of measurements, such as, connectivity measurements which include, for instance, mean degree and degree distributions and correlations. Distances and path lengths as well as hierarchical and spectral measurements are other important categories. Centrality measurements can quantify how important is a node to the network topology for what concerns its robustness and noise tolerance, e.g., betweenness, closeness and eigenvector centrality~\cite{newman2005measure,borgatti2005centrality}. It is also possible to quantify clustering and cycles in a network through measurements like transitivity and the clustering coefficient, which can be used to characterize the small-world property~\cite{watts1998collective}. In addition, Costa \textit{et al.} also addresses the use of classical pattern recognition techniques for network analysis~\cite{da2010pattern}.

%In this paper, we employed a more robust set of measurements and for some specific applications we found out that a combination of different classes of measurements could improve the classification performance for some shape recognition applications.  

In this paper we present an extension of the previous work of Backes \textit{et al.}~\cite{backes2009complex} and we propose a generalized approach for shape characterization in Computer Vision based on network analysis. This approach accounts for different categories of measurements. We have demonstrated how these measurements are related with the properties of shape boundaries. For this task we used samples obtained through an interpolation process performed between regular shapes. This way, we could analyze, for instance, how the emergence of internal angles of a shape influences the connectivity of the resulting network. Besides, we also performed an analysis of the curvature signal of those shapes and its relationship with structural network measurements. We observed that the pixels of high curvature are not the ones with the highest degree, instead, they present a high clustering coefficient.
%Similar analysis was performed for the curvature of those shapes.
In addition, we also evaluated the performance of the proposed approach in three different applications concerning shape recognition. The first image dataset contains geometric shapes of ten different classes. The second dataset contains generic shape contours belonging to nine different categories like animals, fishes and tools among others. This dataset has been used as benchmark in many other studies~\cite{sharvit1998symmetry,sebastian2004recognition}. The last dataset evaluated in this paper is the same dataset yielded by Backes \textit{et al.}, which was used for comparing purposes as well as in order to validate the methodology in a real-world application. This dataset contains images of leaf contours from 30 different plant species. 

This paper is organized as follows: Section~\ref{sec:background} presents a detailed description of the previous work of Backes \textit{et al.} as well as a basic introduction in what concerns the structural characterization of networks. In Section~\ref{sec:methods}, we present a study regarding structural properties of networks for different geometric shapes using an interpolation approach as well as a curvature analysis. In section~\ref{sec:classification}, we evaluated the proposed approach regarding the classification task for three distinct datasets, and, finally, Section~\ref{sec:discussion} presents the discussion.

\section{Background}
\label{sec:background}

\subsection{Previous work}

In the work of Backes~\textit{et al.}~\cite{backes2009complex} is introduced a shape descriptor based on the dynamic evolution of a network built from the contour points. This method is based on a distance criterion in order to establish the connections between the pixels. The threshold parameter, $T$, is used for modeling shape boundaries and to obtain the corresponding feature vectors. All this process is detailed next.

Let $S$ be the contour of an image, such that $S=\{s_1,s_2,\ldots,s_N\}$, where $s_i=[x_i,y_i]$ are the coordinates of point $i$, represented by discrete values. Given a graph of the form $G=\langle V,E \rangle$, each pixel of the contour represents a node in the graph, and, therefore, $S=V$. The set of non-directed edges $E$ is defined for each pair of nodes and the corresponding weight is calculated by the Euclidean distance, as follows:

\begin{equation}
\label{eq:eucDistance}
d(s_i,s_j) = \sqrt{(x_i-x_j)^2 + (y_i-y_j)^2}.
\end{equation}

\noindent The adjacency matrix $W$ of this weighted network is represented by the $N\times N$ matrix, such that $w_{ij} = W([w_i,w_j]) = d(s_i,s_j)$, which is then normalized in the interval $[0,1]$: $W = \frac{W}{\max_{{w_{ij}\in W}}}$. $N$ is the total number of nodes in the network.

Initially, the network is regular, since each node is connected to all the others. At this step, a threshold transformation, $T_l$, can be applied in order to obtain a new set of edges $E'$. This set is composed by the edges whose weights are smaller than $T_l$. If $T_l$ is small, the number of edges will be also small and the network will be composed of many connected components without intersection. Otherwise, if $T_l$ is large, the network will be almost fully connected. For intermediate values of $T_l$, other properties begin appearing, such as the small-world property, characterized by the presence of many connected triples. The authors have demonstrated this property for a set of distinct shapes. Therefore, the threshold parameter controls the connectivity of the network. Notice that the same process can be implemented considering the connections which are above the threshold, i.e., two pixels will be connected if the distance between them is greater than $T_l$.

\begin{figure}[!th]
\centering\includegraphics[scale=0.3]{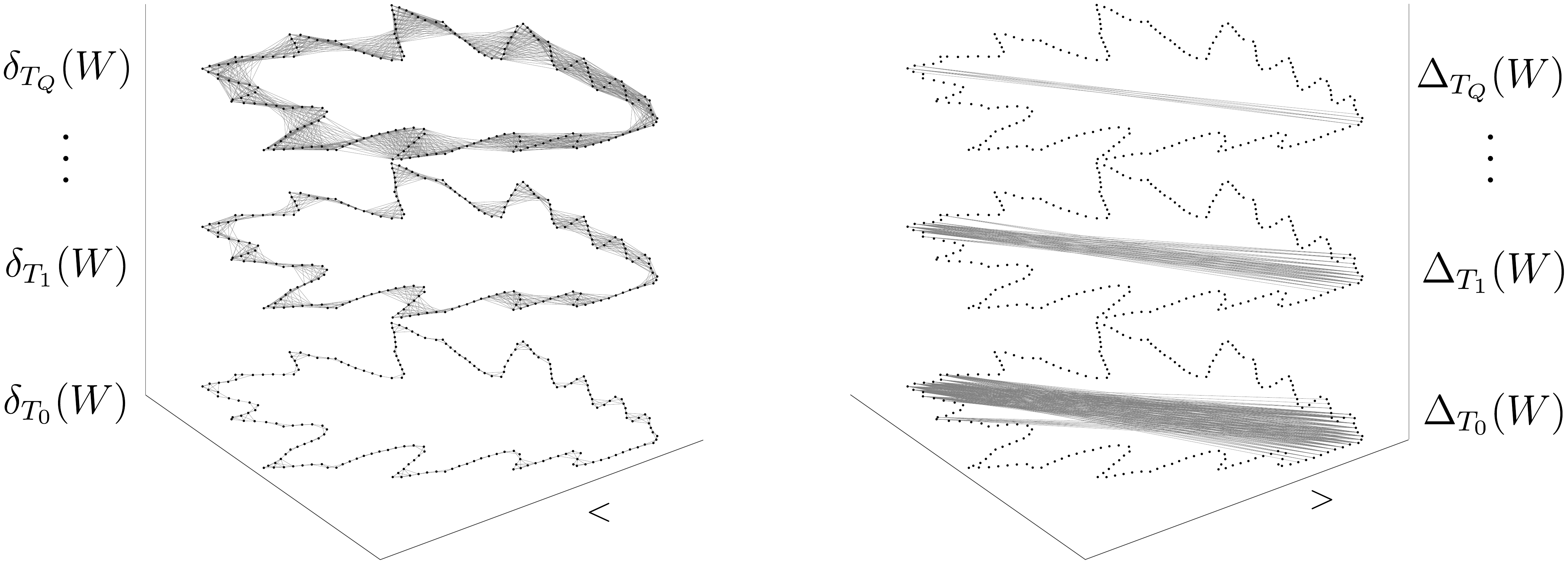}
\caption{Network construction from the contour pixels. The left-side depicts the connections established for different threshold values by applying the comparison \textit{smaller than}. The right-side illustrates the comparison \textit{greater than}.%\textcolor{red}{thresholds on left (0.12, 0.080, 0.04) and right (0.90,0.83, 0.76)
}
\label{fig:abstfig}
\end{figure}

Based on the properties that arise from the transformations defined by $T_l$, the shape descriptor is obtained through the connectivity measures extracted from the network topology. The dynamic evolution of the network as a function of $T_l$ provides distinct attributes which are then combined to compose the feature vector. The transformation is formally defined by the $\delta$ operation as follows:

\begin{equation}
\label{eq:deltaOp}
A_{T_l} = \delta_{T_l}(W) = \forall_W \in W \begin{cases}
                  a_{ij} = 0, & \text{if}~w_{ij} \geq T_l \\
                  a_{ij} = 1, & \text{if}~w_{ij} < T_l
             \end{cases},
\end{equation}

\noindent where $A$ is the resulting unweighted and thresholded matrix. Therefore, the shape characterization is performed through a series of $\delta$ transformations where $T_l$ is regularly incremented by $T_{inc}$. Fig.~\ref{fig:abstfig} illustrates the $\delta_{T_l}(W)$ transformation. The left part of the figure presents the networks obtained by applying the comparison \textit{smaller than}. Therefore, as $T_l$ increases more connections are established. Similar analysis is shown in the right part of the figure, which illustrates the condition \textit{greater than}, $\Delta_{T_l}(W)$, however with the opposite behavior. Given that the distance values are normalized between $0$ and $1$, the threshold values are defined in the same interval. 

%\textcolor{blue}{FIGURA 3D ILUSTRANDO OS THRESHOLDS}

%Many network measures are derived from the connectivity of the nodes represented by their degree, which is the number of neighbors of a node. 
The measurements obtained at each $\delta_{T_l}(W)$ are: the \textit{average degree} ($k_{\mu}$) and the \textit{max degree} ($k_{\kappa}$), which corresponds to the average and to the maximum degree of all the network nodes, respectively. The final feature vector, $\varphi$, is represented by

\begin{equation}
\label{eq:featureVector}
\varphi = [k_{\mu}(T_0),k_{\kappa}(T_0),k_{\mu}(T_1),k_{\kappa}(T_1),\cdots,k_{\mu}(T_Q),k_{\kappa}(T_Q)],
\end{equation}

where the values $k_{\mu}$ and $k_{\kappa}$ correspond to the \textit{average} and \textit{max degree} for each $\delta_{T_l}(W)$ considered. Backes \textit{et al.} \cite{backes2009complex} showed that the proposed descriptor is invariant to rotation and scale transformations applied on images, and, it is robust to noise and image degradation.

In the present paper, in addition to degree-related measurements, we investigated a more generalized approach for shape characterization based on network analysis. Different sets of measurements were evaluated which are capable of describing important structural properties of shape boundaries. In the next subsection, we introduce the definition of some network measurements. And further in this paper we discuss how these measurements can be used as feature vectors.

%, and then, we present the generalized shape descriptor based on the structural properties of the network obtained through the shape contour.

\subsection{Structural network characterization}

A large variety of structural measurements can be used to characterize networks~\cite{costa2007characterization}. It is possible to identify different categories such as connectivity and distance related measurements. Some of these groups are described next.

%We listed seven measurements in this section which were grouped according to common ...

%We employed several topological network measurements in order to assess to the networks properties. The most simplest are regarding the size of a network in terms of the number of nodes $N$. Moreover, the connectivity $k_i$ is the number of neighbors of a vertex $i$ given by $k_i=\sum_{j=1}^{N} A_{ij}$, from which the average degree $\langle k \rangle$ can also be obtained. 

%The degree distribution $P(k)$ represents the probability of finding nodes in a network with degree $k$. Therefore, to compute $\gamma$ we employed a technique based on the accumulated distribution $P(k)$ described in~\cite{Clauset-PowerLawMatlab}. Moreover, measures related to distance such as the average path length, $\langle L \rangle$ and the diameter $D$ were also employed. The average length of the shortest distances between any two nodes of the network is given by $\langle L\rangle= \sum_{i\neq j}\frac{d_{ij}}{N(N-1)}$, which is calculated after computing all pairs of values $d_{ij}$. 

\subsubsection{Connectivity measurements}

The number of neighbors of a given node $i$, $k_i$, is one of the simplest measures regarding the connectivity of a network. From $k_i$ a set of measurements can be derived~\cite{costa2007characterization} as the average degree,

\begin{equation}
\label{eq:avgDegree}
{\langle k \rangle}={\frac{1}{N} \sum_i {k_i}},
\end{equation}

\noindent which is computed considering the degrees of all the nodes in the network. In contrast to $\langle k \rangle$, the hierarchical degree, $k^{h}$, accounts for the connectivity of the node neighbors that are restricted to a hierarchy level $h$. For instance, the hierarchical degree of level 2 of a given node $i$, $k_{i}^2$, is the sum of the degrees of its neighbors. Therefore, the average hierarchical degree of a network, $\langle k^{h} \rangle$, is given by,

\begin{equation}
\label{eq:hierarchDegree}
\langle k^{h} \rangle=\frac{1}{N} \sum_i {k_{i}^h},
\end{equation}

\noindent where $h$ defines the hierarchy level. Another interesting measurement is the clustering coefficient~\cite{watts1998collective}, $cc_{i}$, which express the probability of two vertices $j$ and $k$ being connected to each other since both are connected to node $i$. It is calculated by $cc_i = 2e_i/k_i(k_i-1)$, where $e_i$ represents the number of edges between the neighbors of node $i$. Therefore, the average clustering coefficient, $\langle cc \rangle$, is given by

\begin{equation}
\label{eq:clusteringCoeff}
\langle cc \rangle={\frac{1}{N} \sum_i {cc_i}}.
\end{equation}

\subsubsection{Distance-related measurements}

A \textit{path} that connects two nodes $i$ and $j$ in a network is defined by the sequence of nodes which must be visited to go from $i$ to $j$, and, the number of links is denoted by the distance between them, $d_{ij}$. The geodesic path is the path that has the smallest value of $d_{ij}$. If there is no path between $i$ and $j$, then $d_{ij}=\infty$ or only the largest connected component of the network can be considered. The average path length is defined as follows~\cite{costa2007characterization}:

\begin{equation}
\label{eq:avgPathLength}
{\langle l \rangle}={\frac{1}{N(N-1)} \sum_{i \neq j} {d_{ij}}},
\end{equation}

\noindent where $\frac{1}{N(N-1)}$ is the normalization factor for a totally connected network.

\subsubsection{Degree correlation}

The degree correlation quantifies the tendency of nodes that present high $k$ to connect with nodes that also have a high degree, or, with nodes that present lower values of $k$. It is also called assortativity~\cite{newman2002assortative} and can be calculated as follows:

\begin{equation}
\label{eq:pearsonCorr}
\rho = \frac{{(1/M)\sum_{j>i} {k_i k_j a_{ij}}} - {[(1/M)\sum_{j>i} {(1/2)(k_i+k_j)a_{ij}}]^2}}{{{(1/M)\sum_{j>i} (1/2){(k_i^2+k_j^2) a_{ij}}}} - {{[(1/M)\sum_{j>i} {(1/2)(k_i+k_j)a_{ij}}]^2}}},
\end{equation}

\noindent $\rho$ belongs to the interval $-1 \leq \rho \leq 1$. Value of $\rho$ that are close to $1$ indicate a positive correlation, whereas values close to $-1$, a negative correlation. Finally, values near zero indicate no linear dependency for what concerns the degrees of the network nodes.

\subsubsection{Betweenness centrality}

The betweenness centrality is a measure that accounts for the importance of a node regarding the information load it is responsible in the network~\cite{newman2005measure}. Given $j$ and $k$, two nodes that are not adjacent, i.e., that are not directly linked, the communication between them depends on the nodes that belong to the paths that connect $j$ and $k$. Therefore, the betweenness $b_i$ of a node $i$ is calculated as a function of the number of geodesic paths passing through $i$~\cite{newman2005measure}:

\begin{equation}
\label{eq:betweenness}
b_i = \sum_{j,k,j \neq k} {\frac{n_{jk}(i)}{n_{jk}}},
\end{equation}

\noindent on which, $n_{jk}(i)$ is the number of geodesic paths connecting $j$ and $k$ passing through $i$ and $n_{jk}$ is the number of geodesic paths connecting $j$ and $k$. We adopted the normalized version of Eq.~\ref{eq:betweenness} in the experiments we performed in this paper: $b_i = \frac{1}{n^2}\sum_{j,k,j \neq k} {\frac{n_{jk}(i)}{n_{jk}}}$, where $n^2$ is the total number of possible links. The definition of the betweenness centrality can also be extended to the links of the network, therefore $b_i$ is given as a function of the number of paths passing through an specific link or edge~\cite{costa2007characterization}.

%\subsection{Generalized shape descriptor based on network measurements}
%\section{\textcolor{red}{Shape properties revealed by structural network measurements}}
\section{Shape properties revealed by structural measurements}
\label{sec:methods}

The dynamic evolution of the network, defined by the $\delta_{T_l}(W)$ transformation over the shape boundary, allows the characterization of its contour as well as the intrinsic structural properties that can be associated to that shape. The connectivity of the underlying network is directly influenced by the shape contour, and, therefore, it can be used for pattern recognition.
%investigation of properties related to the connectivity of the network as well as how these properties are influenced by different shapes. 

Besides degree-related measurements, we investigated the performance of a more general set which comprises clustering, distance and centrality measurements. Given that each structural measurement captures specific properties of the network topology, we investigated a set measurements for the characterization of networks which were constructed from regular geometric shapes. The use of these specific shapes allows to relate geometric characteristics, such as angle and curvature, to the structural properties of the network generated according to the proposed approach. 

The next sections illustrate how such measurements are related to shape properties. Therefore, the proposed approach generalizes the method that we presented in the previous section regarding the attributes that can be obtained from the dynamic evolution of the network. From now on, the generalized shape descriptor will be denoted by $\Phi$ in contrast to the network descriptor based only on the degree of the nodes, $\varphi$ (Eq.~\ref{eq:featureVector}).

\subsection{Curvature analysis \& Structural features}

%Shape is one of the most important visual attributes of an object, and it has an important role in shape analysis and representation. However, most of the shape information is redundant and could be discarded without damaging the original shape representation. According to [1], points of high curvature are the most important source of information regarding a shape. The human visual system is capable of recognizing a shape by its higher curvature points. In fact, these points are important features for pattern recognition tasks, as they provide efficient data reduction while retaining crucial shape information [24].

The curvature analysis explores parts of the image that exhibit high contrast and plays an important role in shape recognition. As pointed out in the famous work of Attneave~\cite{attneave1954some}, points that present high curvature are a rich source of information, and the human visual system can recognize a shape considering only those points. 

For the network constructed from the shape boundaries,
%it is expected that pixels with higher curvature values will also present higher mean degree, $\langle k \rangle$, since those pixels are the most connected ones. Accordingly, this is also observed for the hierarchical degree, $\langle H_{k_m} \rangle$. 
it should be noticed that, the degree of a pixel is also conditioned to the threshold parameter $T_l$. Therefore, the smaller the internal angles of a shape, the smaller the distance between the pixels of the shape boundary and, consequently, the higher the mean degree. 
%\textcolor{red}{same behavior can be observed for the mean clustering coefficient $\langle cc \rangle$. Small distances between pairs of pixels favors the formation of small-world graphs.}
% * <gisele.hbm@gmail.com> 2017-10-24T16:31:05.712Z:
%
% ^.
% * <gisele.hbm@gmail.com> 2017-10-24T16:30:54.756Z:
%
% > The curvature analysis explores parts of the image which exhibit high contrast and plays an important role in shape recognition. As pointed out by Attneave~\cite{attneave1954some}, points of an image that present high curvature are a rich source of information, and the human visual system can recognize a shape only through these points. 
% > For the network constructed from the shape boundaries,
% > %it is expected that pixels with higher curvature values will also present higher mean degree, $\langle k \rangle$, since those pixels are the most connected ones. Accordingly, this is also observed for the hierarchical degree, $\langle H_{k_m} \rangle$. 
% > it should be noticed that, the degree of a pixel is also conditioned to the threshold parameter $T_l$. Therefore, the smaller the internal angles of a shape, the smaller the distance between the pixels of the shape boundary and, consequently, the higher the mean degree. 
% > %\textcolor{red}{same behavior can be observed for the mean clustering coefficient $\langle cc \rangle$. Small distances between pairs of pixels favors the formation of small-world graphs.}
%
% ^.

\begin{figure}[!ht]
\centering\includegraphics[scale=0.38]{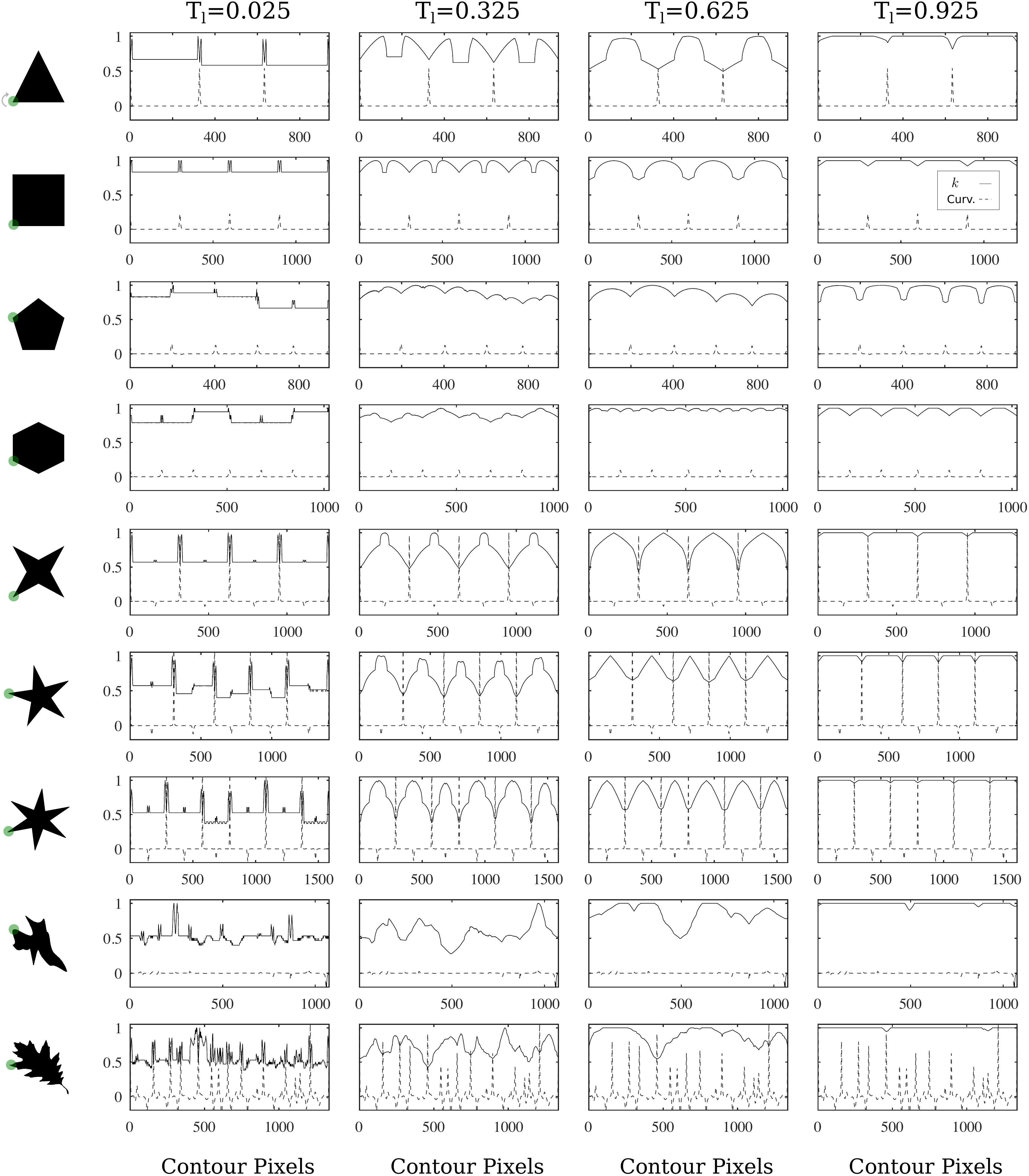}
\caption{Comparison of the curvature signal (dashed line) and the degree of each node as a function of the threshold parameter $T_l$ for different geometric shapes.}
\label{fig:formas-curvas}
\end{figure}

\begin{figure}[!ht]
\centering\includegraphics[scale=0.38]{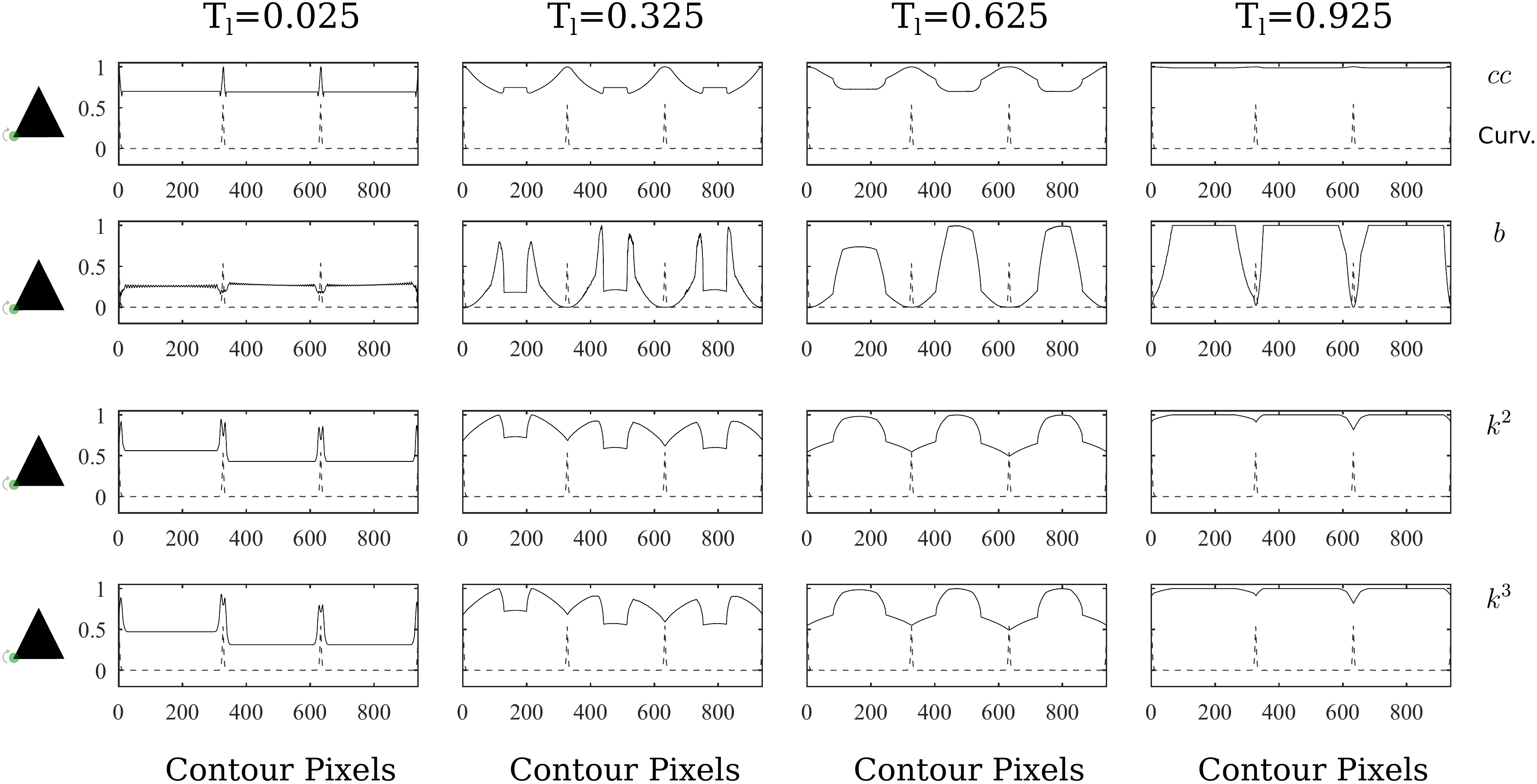}
\caption{Comparison of the curvature signal (dashed line) and other structural measurements: clustering coefficient, betweenness, hierarchical degree of level 2 and  hierarchical degree of level 3 (from top to bottom), as a function of the threshold parameter $T_l$ for the triangular shape.}
\label{fig:curvaturaOtros}
\end{figure}

Fig.~\ref{fig:formas-curvas} illustrates how curvature $C$ (dashed line) and the node degree are related to each other for different shapes. The values of both measurements were normalized between 0 and 1. Each row represents a specific shape and each column represents a different threshold which increases from left to right. The $x$-axis of each plot corresponds to the pixels of each shape, therefore, the curvature and the other measurements are presented per pixel. The method used for the estimation of the curvature signal is based on the DFT (\textit{Discrete Fourier Transform})~\cite{cesar1996towards}. The initial pixel (point), from which the curvature and is calculated, is highlighted for each polygon as a green dot. The aliasing effect is more apparent for some shapes, like the hexagon. This effect is reflected as small oscillations in the curvature, however we can easily identify the points of maximum curvature which are close to the corners of those shapes. For the square shape (second line), for instance, there are four equally spaced peaks that correspond to the corners. In general, for all shapes, it is possible to observe that, except for $T_l=0.025$, pixels with the highest curvature are the ones with the lowest degrees. This occurs because they are distant from the other pixels when compared to the points located on the sides of the square. For $T_l=0.025$, the most connected pixels are the ones near the corners, since at this level the network presents only a few connections. Then, as the threshold increases and the network gets more connections, the pixels between two corners are the ones with the highest degrees.
%It is also possible to observe that the pixels that present the highest curvature values are also the pixels with the highest mean degree (\textcolor{red}{cor da curva}) for all the thresholds. 
For what concerns the stars, we can observe a complementary behavior between the corners with positive and negative curvature. Finally, the last two rows present the degree and curvature signals belonging to irregular shapes, for which the general assumptions are also valid.

%The other geometric shapes present the same properties highlighted in the last paragraph. The curvature analysis for the square can be applied to the triangle. The curvatures of the pentagon and the hexagon were more influenced by the aliasing effect, mainly for the threshold $T_l=0.025$. Regarding the stars, it can be observed the complementary behavior between the convex and the concave corners (negative curvature). Finally, the two last rows present the behavior or two irregular shapes, a leaf and an abstract shape.

Similar comparison was performed for other measurements which were also obtained from the network structure. The results are presented in Fig.~\ref{fig:curvaturaOtros} for the triangular shape. An opposite behavior was obtained for the clustering coefficient curve (first row) when compared to the degree analysis presented in Fig.~\ref{fig:formas-curvas}. The pixels with the highest curvature are the ones with the highest values of $cc$. In the case of the triangle, the corner presents connections between the two line segments that define this corner, which increases the clustering coefficient of that node. Moreover, in general, for the great majority of the plots, all the analyzed shapes present high values of average clustering coefficient (greater than 0.5). As the threshold increases, the number of connected triples of the corresponding network also increases, but, now, also reflecting the increase for what concerns the density of the network nodes. This means that if pixel $i$ is connected to pixel $i-1$, given a radius $r$, then pixel $i+1$ will also be connected to pixel $i-1$, giving rise to a small-world network. 

For what concerns the betweenness centrality (second row of Fig.~\ref{fig:curvaturaOtros}), it is possible to observe that, similarly to the node degree, that measurement is also high for the nodes that are not in the corners of the shape. As both $T_l$ and the number of connections of the network increase, new pathways are established among the pixels. Consequently, the pixels with the highest curvature values will not be in the shortest paths, which influences the betweenness of the node. Finally, the last two rows of Fig.~\ref{fig:curvaturaOtros} illustrate the relationship between the curvature and the hierarchical degree of levels two and three, respectively, for each pixel. Once more, we can observe that the general behavior of both measurements follows the same trend observed for the node degree, as observed in Fig.~\ref{fig:formas-curvas}.

Moreover, both Figs.~\ref{fig:formas-curvas} and \ref{fig:curvaturaOtros} illustrate the behavior of the dynamical properties of the networks built upon shape contours. The calibration of $T_l$ may give rise to sparsely or highly connected networks, as well as other intermediate connectivity patterns. For all the plots presented in these figures we used the comparison \textit{smaller than}, but the same analysis can be performed for the comparison \textit{greater than}. Together with the curvature analysis, we demonstrated the properties of the networks evolved on the topologies defined over the contour pixels.

%\textcolor{red}{ADD SOME FINAL CONCLUSION / DISCUSSION ABOUT THESE FIGURES LINKING THEN TO ATTRIBUTES OF NETWORKS THAT CAN BE USED FOR SHAPE RECOGNITION, LIKE A SIGNAL MEASURED BY TIME STEP, SIGNATURE} 

%In spite of being highly connected, the pixels of the shape corner are not the ones with the highest values of $\langle cc \rangle$, although, in general, for the great majority of the plots, all the shapes already present high values of average clustering coefficient (greater than 0.5). In such a regular shape, the pixels of the square side present regular neighborhoods, which increases the number of connected triples of the corresponding graph, e.g., if pixel $i$ is connected to pixel $i-1$, given a radius of 2, then pixel $i+1$ will also be connected to pixel $i-1$, giving birth to the small-world graph. However, the corner pixels have neighbors in two sides of the square, but some of these neighbors are not connected as triples which decreases the values of their clustering coefficients. For $T_l=0.025$, it is easier to visualize such properties. As $T_l$ increases, the same behavior is observed, but now also reflecting the increase in the density of the network nodes.

\subsection{Shape Interpolation Analysis}
\label{S:3}

\begin{figure}[!th]
\centering\includegraphics[scale=0.4]{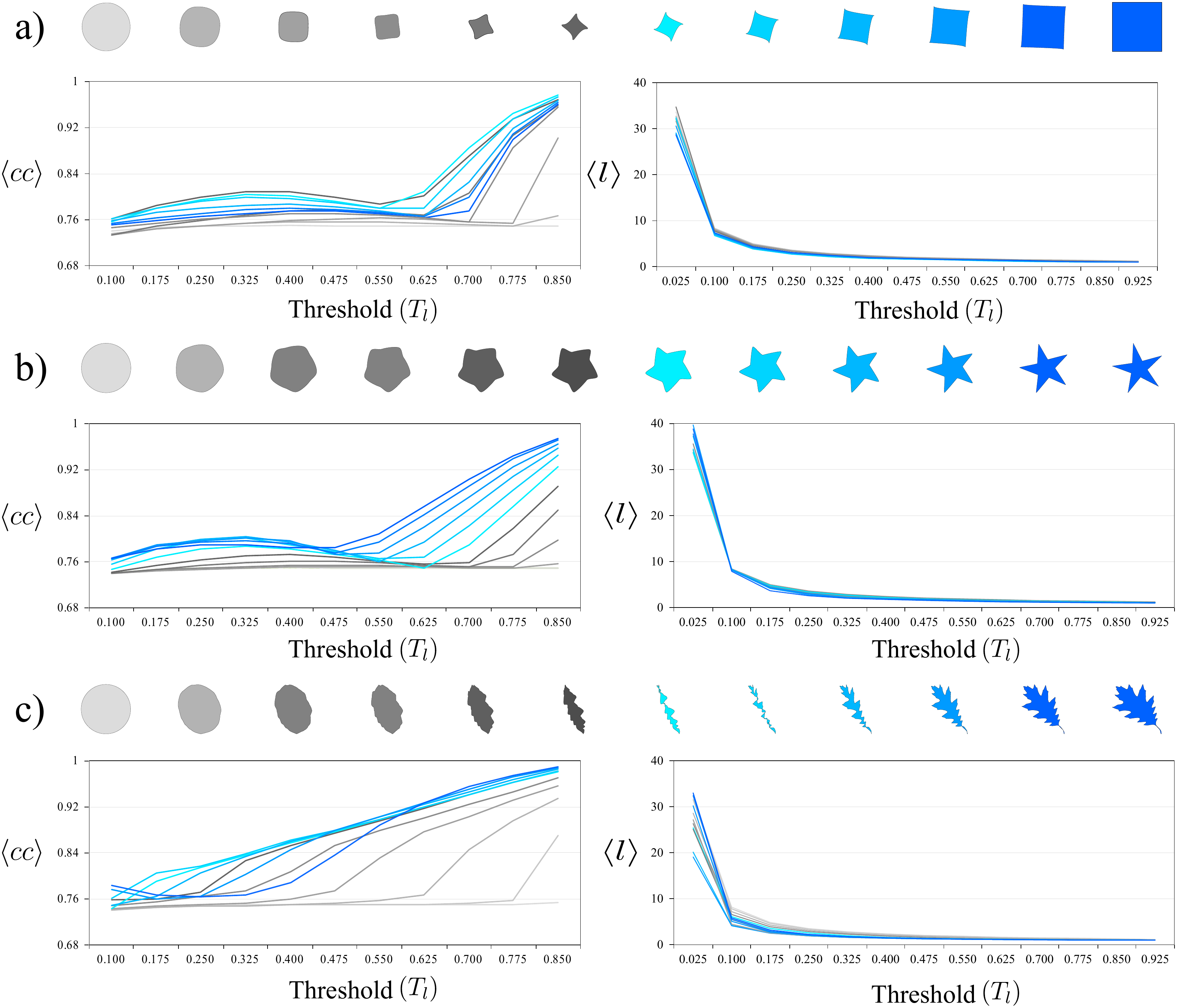}
\caption{Interpolation analysis. Each row represents the interpolation between two shapes. The plots in the left side show the clustering coefficient as a function of the threshold parameter, $T_l$, and, the plots in the right side show the average path length also as a function of $T_l$. The colors make the correspondence with the shapes.}
\label{fig:interpolation}
\end{figure}

%Given that each structural measurement captures specific properties of the network topology, we investigated a set measurements for the characterization of networks which were constructed from regular geometric shapes. The use of these specific shapes allows to relate geometric characteristics, such as angle and curvature, to the structural properties of the network generated according to the proposed approach. 
%We evaluated networks obtained for triangular, squared, circular, hexagonal and other geometric shapes.
%The first rows of figures Fig.~\ref{fig:interpolation}-a), b) and c) 
Fig.~\ref{fig:interpolation} presents three interpolation processes that consist of different series of shapes representing the steps performed to fill the space between two geometric shapes. Particularly, the first and the last shapes of the sequence are the reference shapes 
%(\textcolor{red}{TALVEZ AQUI FALAR MAIS FORMALMENTE DO PROCESSO DE INTERPOLAC\~{A}O... VERIFICAR NO INKSCAPE QUAL METODO ELE USA}).

Fig.~\ref{fig:interpolation}-a) presents the interpolation between a circle and a square. The two plots of this figure correspond to the average clustering coefficient $\langle cc \rangle$ and the average path length $\langle l \rangle$, as a function of the threshold parameter, $T_l$. Each color connects the geometric shape to its corresponding measurements which were obtained for different values of $T_l$. We can observe in the interpolation process the emergence of internal angles, and, consequently, the curvature values for some pixels start to increase.
%, and, consequently, an increase in the curvature values for each specific shape. 
As largely explored in literature~\cite{watts1998collective,costa2007characterization}, small-world networks are characterized by a high clustering coefficient and a small average path length. As expected, independently of $T_l$, the value of $\langle cc \rangle$ is constant for the circular shape (light gray curve). Regarding the square (dark blue curve), the values of $\langle cc \rangle$ tend to increase with the threshold $T_l$. This is observed due to the fact that the mean degree, $\langle k \rangle$, also increases with $T_l$, therefore, more connections will be established among the contour, and the clustering coefficient will increase.

Similarly, Fig.~\ref{fig:interpolation}-b) presents the interpolation between a circle and a star. Since the star has acute angles, the values of $\langle cc \rangle$ are higher from $T_l > 0.55$ when compared to the corresponding plot of Fig.~\ref{fig:interpolation}-a). The interpolation process also introduces some irregularities in the intermediate shapes, therefore, they are not totally uniform as the circle and the square. Consequently, the values of $\langle cc \rangle$ for these irregular shapes tend to be higher. Finally, Fig.~\ref{fig:interpolation}-c) presents the interpolation between two irregular shapes. A rapid growth in the values of $\langle cc \rangle$ is observed also due to the irregularities of the shapes. 

Meanwhile, the analysis of the average path length for the three interpolations indicates no significant differences for what concerns the different thresholds. It is possible to observe that $\langle l \rangle$ is high for thresholds $T_l < 0.1$ and present a fast decay. This can be explained by the fact that for such small thresholds the network is not totally connected and, therefore, the path between two nodes is large or that it may not exist. In order to calculate $\langle l \rangle$, we consider the missing paths as having distance one value higher than the largest possible geodesic path.

\begin{figure}[t]
\centering\includegraphics[scale=0.65]{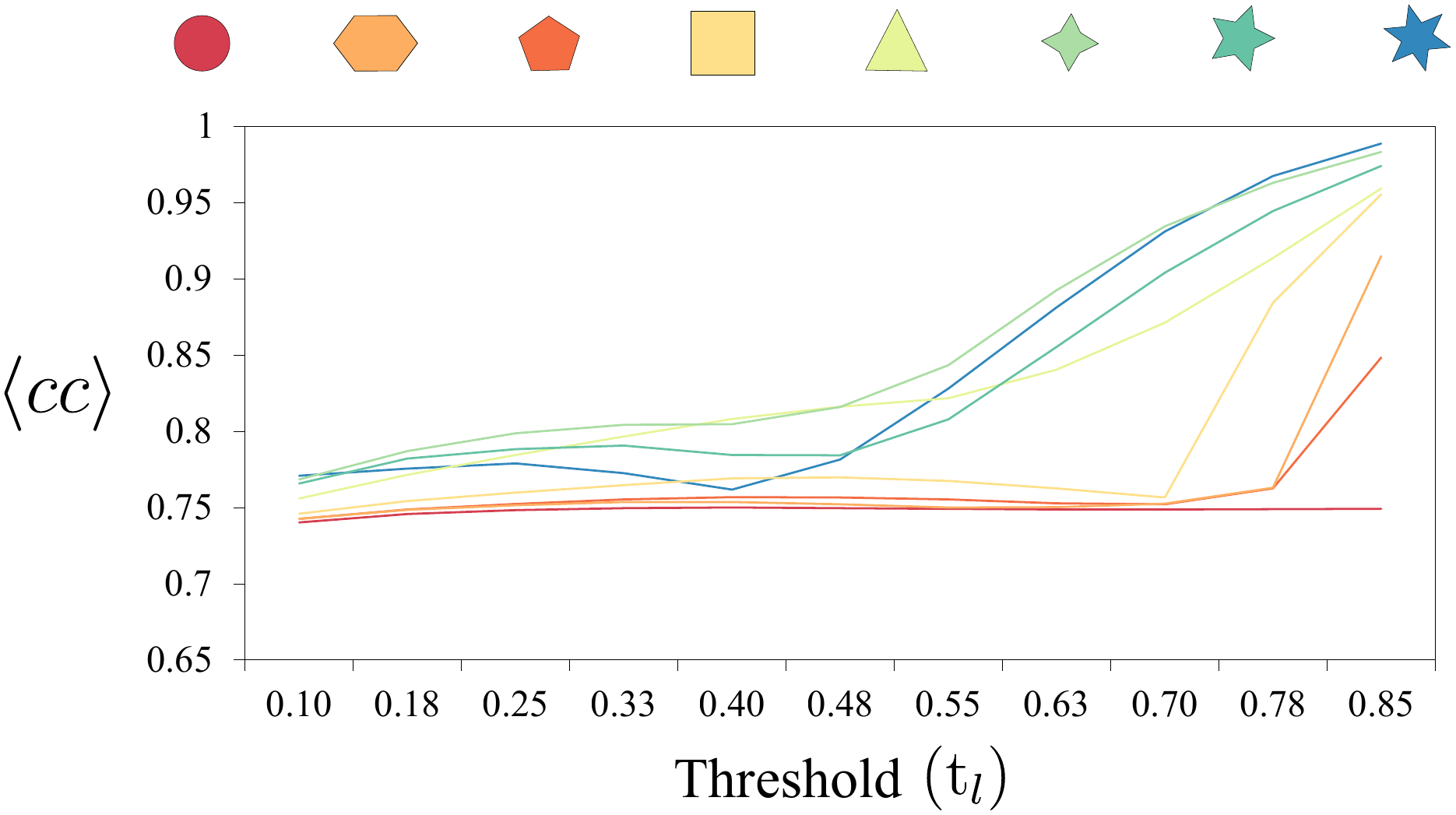}
\caption{Average clustering coefficient ($\langle cc \rangle$) as a function of the threshold parameter $T_l$ for eight geometric shapes. Each color corresponds to a specific geometric shape.}
\label{fig:interpolationclustering}
\end{figure}

We also performed an analysis of the clustering coefficient as a function of the internal angles of different geometric shapes, which can be seen in Fig.~\ref{fig:interpolationclustering}. From left to right the internal angles decrease and the respective average clustering coefficients increase rapidly, since the distance between the pixels will decrease and more connections will be established between them.

%\textcolor{red}{The number of pixels also increases from left to right, and, therefore, the distance between pixels will decrease and more connections will be established between them. Should be noticed that, due to the aliasing sampling, for more irregular shapes the number of pixels will increase and therefore...}

%In the next section we introduce some shape classification problems which illustrate how the different measurements studied in this section can be used to compose a feature vector, and, consequently, can be used for pattern recognition.

\section{Shape classification}
\label{sec:classification}

In this section we present an analysis of the structural network measurements when used as feature vectors for shape classification. We evaluated different image datasets regarding the correct identification of the categories of interest. This evaluation was performed taking into account both the threshold variations as well as the different combinations of structural measurements chosen to compose the feature vector. In addition, we also evaluated the classification accuracy in the presence of noise for what concerns the shape structure and when applying geometric transformations as rotation and scale.   

\subsection{Datasets}
\label{sec:datasets}

\begin{figure}[!ht]
\centering\includegraphics[scale=0.48]{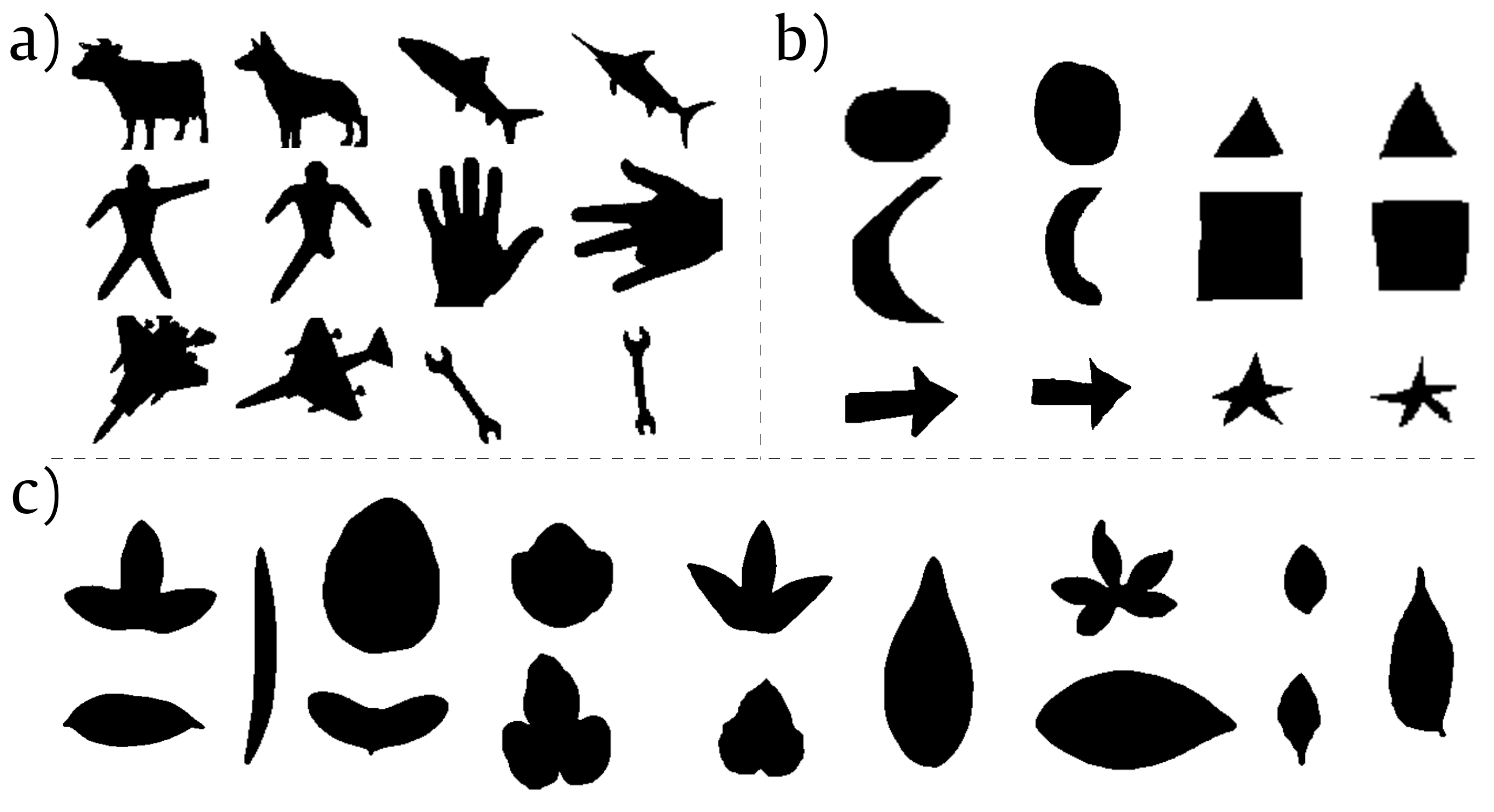}
\caption{Sample instances of the three datasets evaluated in this paper. a) \textit{geom-shapes}, b) \textit{generic-shapes} and c) \textit{leaves}.}
\label{fig:dataset}
\end{figure}

Three image datasets were used in the experiments performed in this paper. The first dataset, the \textit{geom-shapes} dataset, is composed of 250 geometric shapes of 10 distinct classes: circle, triangle, square, pentagon, hexagon, star, arrow, ``pacman'', heart and moon. All these shapes were manually designed by different people. Therefore, the shapes of the same class contain intrinsic variations due to the free-hand design. Also the shapes of this dataset incorporate variations regarding the size and the structure of the geometric objects. Samples of this dataset are shown in Fig.~\ref{fig:dataset}-a). 

The second dataset, the \textit{generic-shapes}, contains generic shapes from different categories, which also present structural variations among instances of the same class. The \textit{generic-shapes} dataset is composed by 99 generic shapes of 9 different classes~\cite{sharvit1998symmetry,sebastian2004recognition}. This dataset has been used as a benchmark comparison in many computer vision applications. Some examples of shapes belonging to this dataset can be seen in Fig.~\ref{fig:dataset}-b). 

Finally, the last dataset was yielded by Backes~\textit{et al.}~\cite{backes2009complex} and contains real-world leaf contours from distinct plants. The \textit{leaves} dataset is composed by 600 images of leaf contours. There are 600 images from 30 distinct classes of plants. Therefore, there are 20 image samples for each class. The leaf images comprise distinct trees of the Brazilian vegetation. The complete description of this dataset is available in~\cite{leavesDataset}. Fig.~\ref{fig:dataset}-c) presents some image samples belonging to the \textit{leaves} dataset.
All the three datasets presented in this section have uniform distribution of classes, i.e., all the classes of each dataset have the same number of instances. The composition of each dataset is summarized next.

\subsection{Training and validation strategies}

We used \textit{n}-fold cross-validation to evaluate the performance of the shape classification tasks for the different datasets. Therefore, for each validation, we have different training and validation sets. The cross-validation procedure was applied 100 times for each dataset and the standard deviation was calculated over all these runs since the instances are randomly assigned to each fold. We analyzed the classification performance for three classifiers: $k$-NN (\textit{$k$-Nearest Neighbors}) for $k=1$, SMO (an optimized version of SVM - \textit{Support Vector Machines})~\cite{Platt1998,Keerthi2001} and Naive Bayes. 

The performance of the classifier is given by the number of correctly classified instances of a dataset in relation to the total number of instances, i.e., the accuracy represents the percentage of correctly classified instances. Finally, the measurements chosen to compose the feature vector are: $\langle k \rangle$ (mean degree), $\langle k^h \rangle$ (average hierarchical degree), $\langle cc \rangle$ (clustering coefficient), $\langle l \rangle$ (average path length), $\rho$ (degree correlation) and $\langle b \rangle$ (average betweenness).
%These measurements were evaluated individually and also combined in the feature vector.

\subsection{Performance evaluation and comparison with CNDescriptor}
%\subsection{Performance evaluation}
%\subsection{Performance results}

\begin{figure}[!ht]
\centering\includegraphics[scale=0.23]{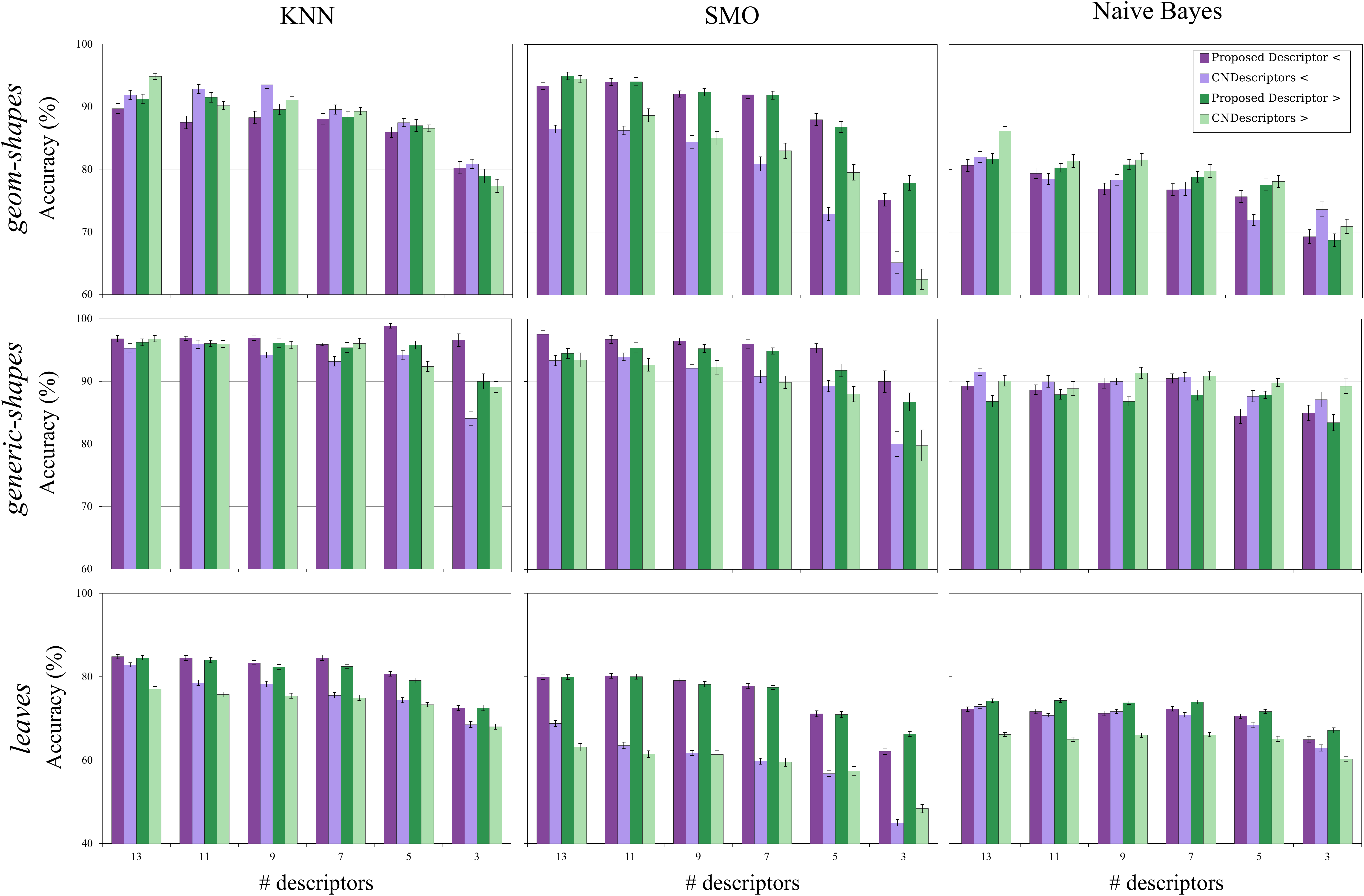}
\caption{Classification accuracy (\%) and standard deviation obtained for the datasets a) \textit{geom-shapes}, b) \textit{generic-shapes} and c) \textit{leaves}, using the proposed shape descriptor ($\Phi$) and different classifiers.}
\label{fig:accuracy}
\end{figure}

The performance of the proposed shape descriptor was evaluated when applied to shape classification tasks considering the three datasets described in the previous section: \textit{geom-shapes}, \textit{generic-shapes} and \textit{leaves}. In Fig.~\ref{fig:accuracy} we can observe the classification performance (\%) for each dataset and for each classifier, with the corresponding standard deviation. The different colors, purple and green, represent the two different configuration settings of the descriptor: connections (network links) \textit{smaller than} or \textit{greater than} the threshold $T_l$. Meanwhile, the different shadings of each color represent the comparison between both network-based descriptors: the generalized ($\Phi$) and the degree specific ($\varphi$). For each single plot, the different bar groups on the \textit{x}-axis, correspond to the number of thresholds resulting from the sampling of the threshold range ($[0,1]$), i.e., the different number of thresholds, $n_T$, chosen to compose the feature vector for a specific run of the experiment. For instance, the first bar group corresponds to 13 values of thresholds equally spaced in the range $[0,1]$. The $n_T$ times the number of network measurements correspond to the final number of attributes. In this experiment, all the network measurements were combined in the same feature vector: $\Phi = [\langle k \rangle(T_0), \allowbreak \langle k^2 \rangle(T_0), \allowbreak \langle k^3 \rangle(T_0), \allowbreak \langle cc \rangle(T_0), \allowbreak \langle l \rangle(T_0), \allowbreak \rho(T_0), \allowbreak \langle b \rangle(T_0), \allowbreak \ldots, \allowbreak \langle k \rangle(T_{n_T}), \allowbreak \langle k^2 \rangle(T_{n_T}), \allowbreak \langle k^3 \rangle(T_{n_T}), \allowbreak \langle cc \rangle(T_{n_T}), \allowbreak \langle l \rangle(T_{n_T}), \allowbreak \rho(T_{n_T}), \allowbreak \langle b \rangle(T_{n_T}) ]$.

An analysis of Fig.~\ref{fig:accuracy} reveals that the performances of all the classifiers decrease when using a small number of attributes. However, there is a convergence in what concerns the accuracy values when $n_T$ is increased. This result shows the influence of the number of attributes, and, consequently, the influence of the sampling size of the threshold interval in the final performance. Regarding the comparison between the two configuration settings of the descriptor for the threshold $T_l$, no significant differences in accuracy were observed. This result leads to the conclusion that both \textit{smaller than} and \textit{greater than} approaches provided similar performances, except in a few cases, like the \textit{barplot} showing the \textit{Naive Bayes} performance for the \textit{geom-shapes} dataset. In contrast, we can observe very similar accuracy values for both settings in the SMO \textit{barplot} for the \textit{leaves} dataset.

For the \textit{geom-shapes} dataset, using a larger set of measurements only improved the classification accuracy for SMO classifier. In this case, the feature vector, $\Phi$, provided a significant increase in the final performance when compared to $\varphi$. However, this improvement was not observed for the other two classifiers and, in this case, $\varphi$ provided better results for many different values of $n_T$. This can be observed for both comparisons, \textit{smaller than} and \textit{greater than}. Notwithstanding, from an overall analysis of the results for this dataset, it is possible to observe that the difference in performance between the two feature vectors is small and their use is interchangeably in many cases. The same analysis can be applied for the results obtained for the \textit{generic-shapes} dataset, even with a smaller difference in accuracy when comparing both feature vectors. For k-NN classifier, for instance, the performance improvement when using a more robust set of measurements is more evident for smaller values of $n_T$. This is mainly related to the fact that when there is a greater variability of the measurements that are in the feature vector, the sampling of the threshold interval can be reduced.

For what concerns the \textit{leaves} dataset, there was a considerable improvement in the classification performance when using the proposed feature vector, $\Phi$. Its robustness to the threshold sampling can also be observed, specially for the SMO classifier. This result also accounts for the fact that when using a reduced $n_T$, a greater variability of measurements in the feature vector is more suitable and computationally more efficient, since the vector dimension is also reduced. A major difference between the \textit{leaves} and the other two datasets is that the first represents real-world shapes. Also, it is a challenging dataset regarding the number of classes (30) and the number of samples per class (20). The \textit{geom-shapes} dataset, for instance, was evaluated mainly as a benchmark comparison. Therefore, in this case it is expected high accuracy values for both feature vectors.

\subsection{Single threshold analysis}

\begin{figure}[!ht]
\centering\includegraphics[scale=0.38]{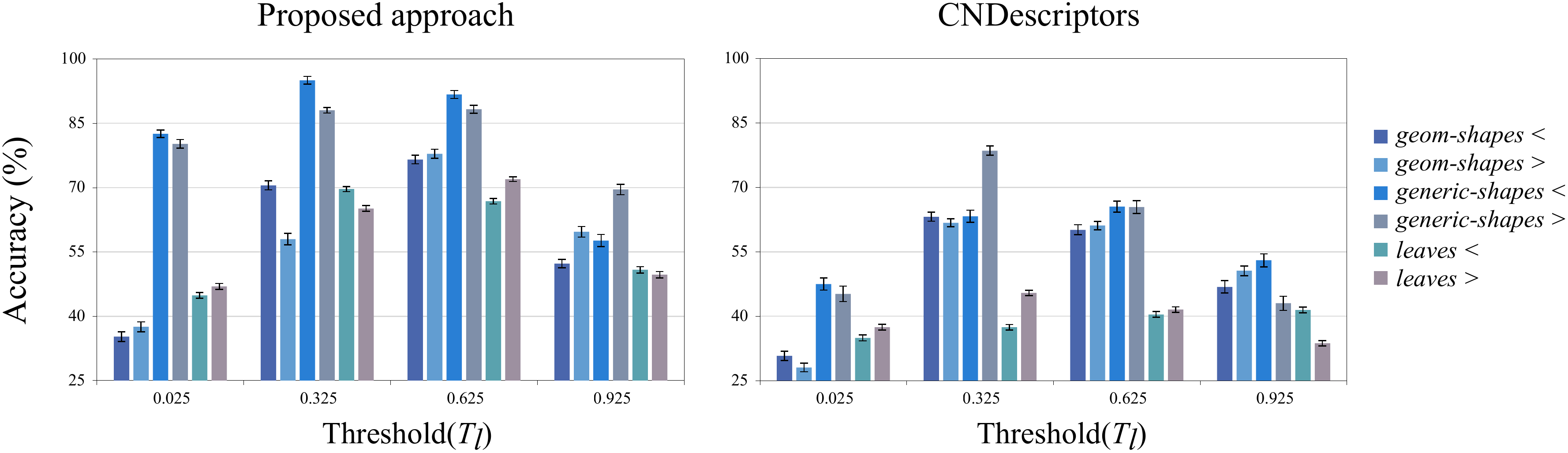}
\caption{Classification accuracy (\%) and standard deviation obtained for the datasets a) \textit{geom-shapes}, b) \textit{generic-shapes} and c) \textit{leaves}, using the proposed shape descriptor ($\Phi$) for a single threshold value $T_i$ without considering its dynamical evolution.}
\label{fig:singleThreshold}
\end{figure}

A main characteristic of the proposed methodology is the dynamic evolution of the network constructed over the shape contour. This evolution is implemented as a subsequent extraction of measurements from the network topology for each incremental threshold. A small $T_l$ will result in a poorly connected network and, in contrast, a high $T_l$ will provide a highly connected network. Therefore, the classification accuracy is improved with the union of the attributes obtained incrementally. However, it is expected in such a method that the most extreme thresholds will contribute less to the final accuracy than the intermediate thresholds. In the experiments presented in this section, we evaluated the accuracy provided by the different threshold values individually. The result for each dataset and for each configurations of the shape descriptor (\textit{smaller than} and \textit{greater than}) are presented in Fig.~\ref{fig:singleThreshold}.

As expected the extreme thresholds are less accurate in both plots. In spite of that, the performance obtained when using $\Phi$ descriptor is much higher than the performance obtained with the $\varphi$, for all the datasets. This result corroborates the previous experiments when a small threshold sampling was used. In this case, a large variability regarding the measurements can increase the classification performance. Another result that should be highlighted in this experiment is the fact that for both descriptors the concatenation of network measurements for a sequence of thresholds, taking into account its dynamic evolution, will provide better accuracies than considering only a single threshold.

\subsection{Robustness \& noise tolerance analysis for the \textit{leaves-dataset}}

In the work of Backes~\textit{et al.}~\cite{backes2009complex}, the authors have shown that the CNDescriptor is invariant to scale and rotation transformations and also that it is robust to noise and image degradation. Within this context, the same characteristics were observed using the proposed approach in this paper which generalizes the feature vector regarding the measurements extracted from the network topology ($\Phi$). We evaluated how such measurements as well as the classification accuracy are affected when those transformations are applied to the \textit{leaves} dataset.

\begin{figure}[!ht]
\centering\includegraphics[scale=0.3]{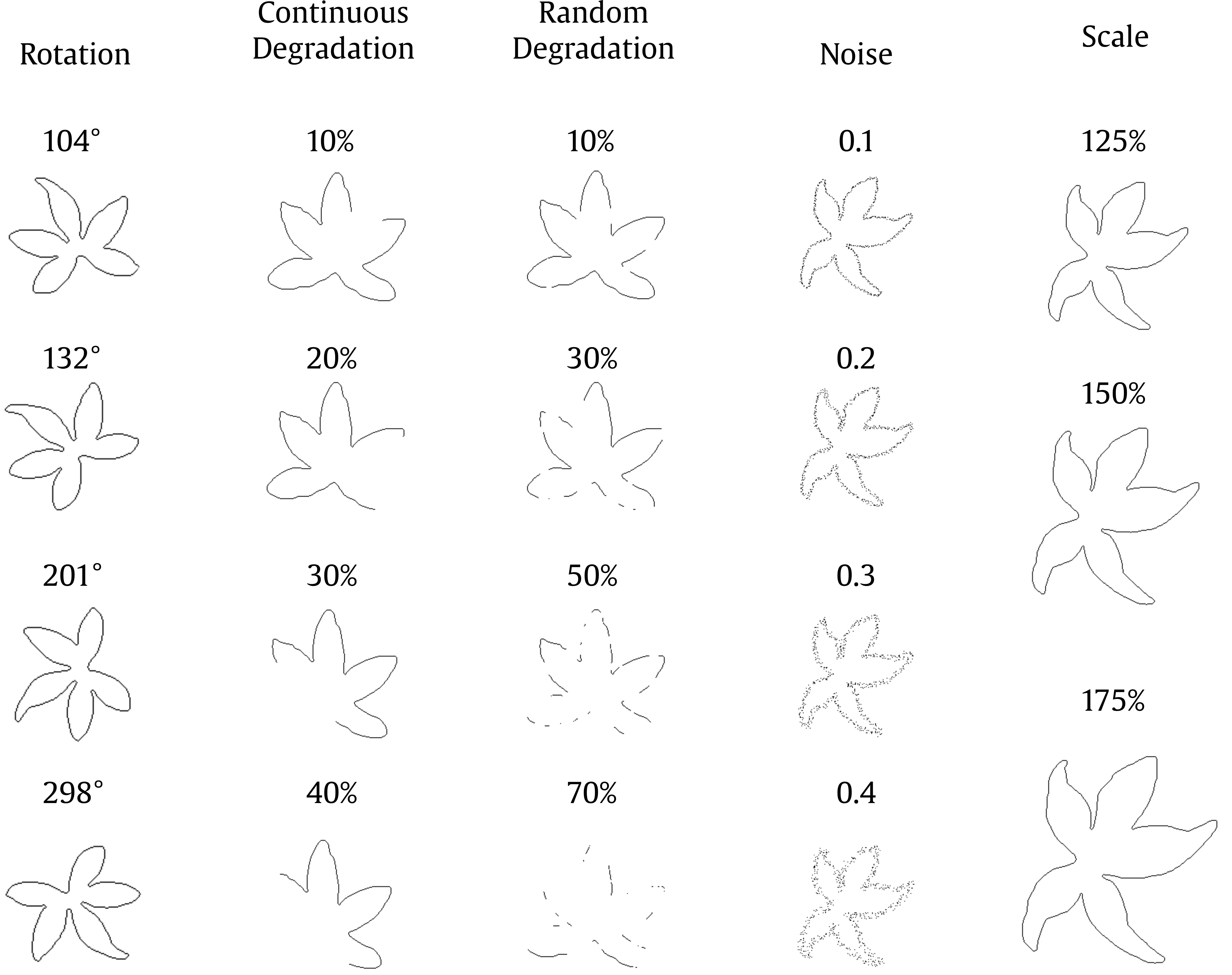}
\caption{Transformations applied to the images of the \textit{leaves} dataset (per column): 1) rotation by the angles 104$^{\circ}$, 132$^{\circ}$, 201$^{\circ}$ and 298$^{\circ}$; 2) Degradation (continuous and random) on which parts of the contour are removed; 3) Noise, which was uniformly generated, and, 4) Scale by the factors: 125\%, 150\% and 175\%.}
\label{fig:datasetPerturbed}
\end{figure}

The first transformations are rotation and scale. The \textit{leaves-rotated} dataset contains the original images of the \textit{leaves} dataset rotated by the following angles: 7$^{\circ}$, 35$^{\circ}$, 104$^{\circ}$, 132$^{\circ}$, 201$^{\circ}$ and 298$^{\circ}$, and, the \textit{leaves-scaled} dataset is composed by the same original images scaled by the factors: 200\%, 175\%, 150\% and 125\%. The noise tolerance and the robustness to degradation are other two properties that we investigated for the \textit{leaves} dataset. The noise applied to the original images was uniformly generated in the range $[-n \ldots n]$, being $n$ the intensity level of the noise. Given the bi-dimensional contours of the shapes, the noise pattern is applied in the $x$ and $y$ coordinates. The \textit{leaves-noise} dataset contains four different noise values applied to each original image. Finally, the robustness to degradation property was evaluated using the \textit{leaves-degraded} dataset, which contains 17 different levels of degradation which were applied in each leaf image. This dataset presents two types of degradations, continuous and random. In the continuous degradation, only adjacent pixels of the shape contour are eliminated. Meanwhile, in the second type of degradation the pixels are randomly chosen to be eliminated. For both methods, as higher the degradation level is, more pixels will be removed from the contour. Fig.~\ref{fig:datasetPerturbed} presents some examples of the transformations applied to the \textit{leaves} dataset. 

\newpage
\begin{figure}[!ht]
\centering\includegraphics[scale=0.9]{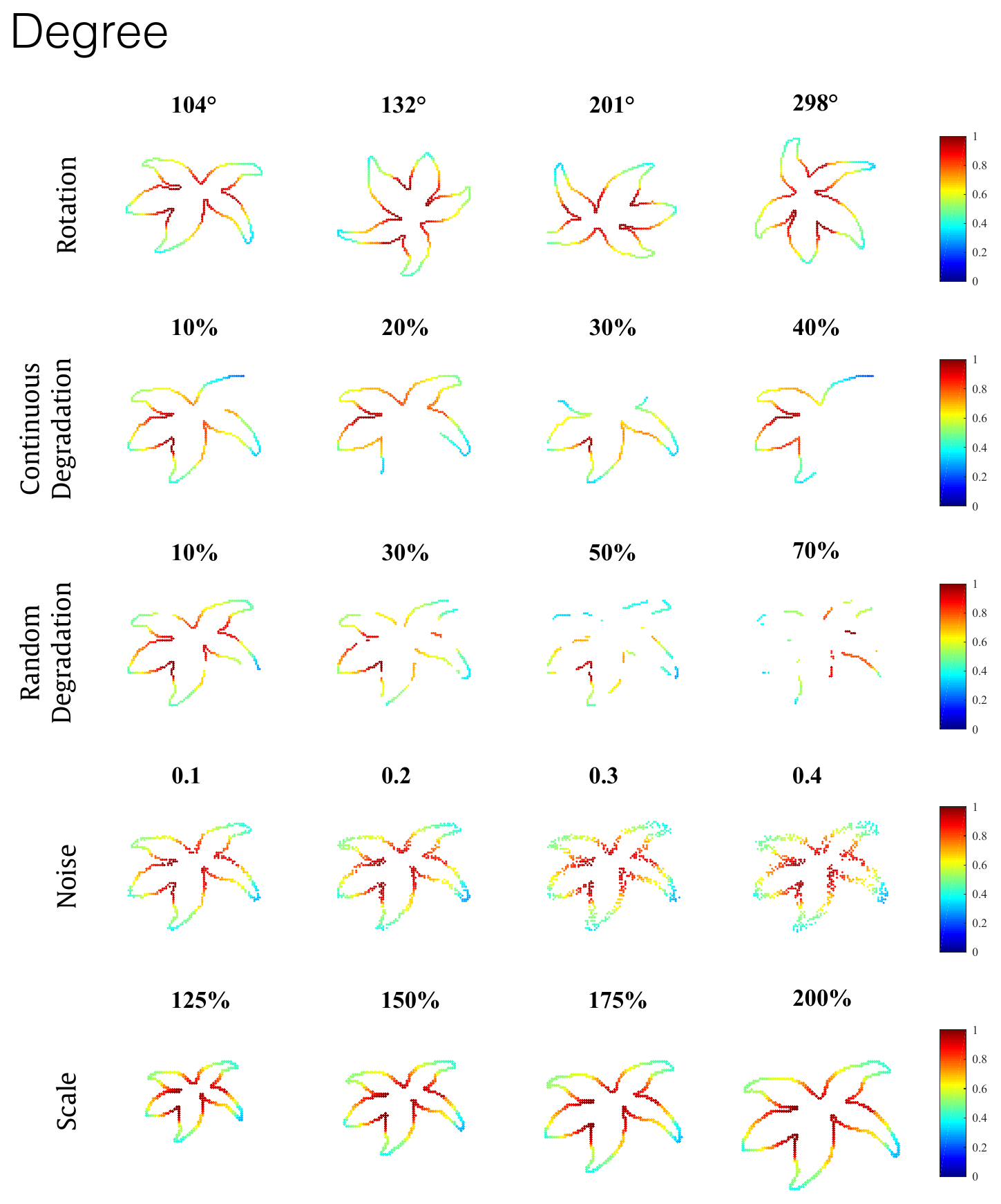}
\caption{Color-map representing the degree ($k$) of the contour pixels using threshold $T_l=0.325$ for the network construction. Each row illustrates a transformation and its respective parameters.}
\label{fig:degreeHeatmap}
\end{figure}

\newpage
\begin{figure}[!ht]
\centering\includegraphics[scale=0.9]{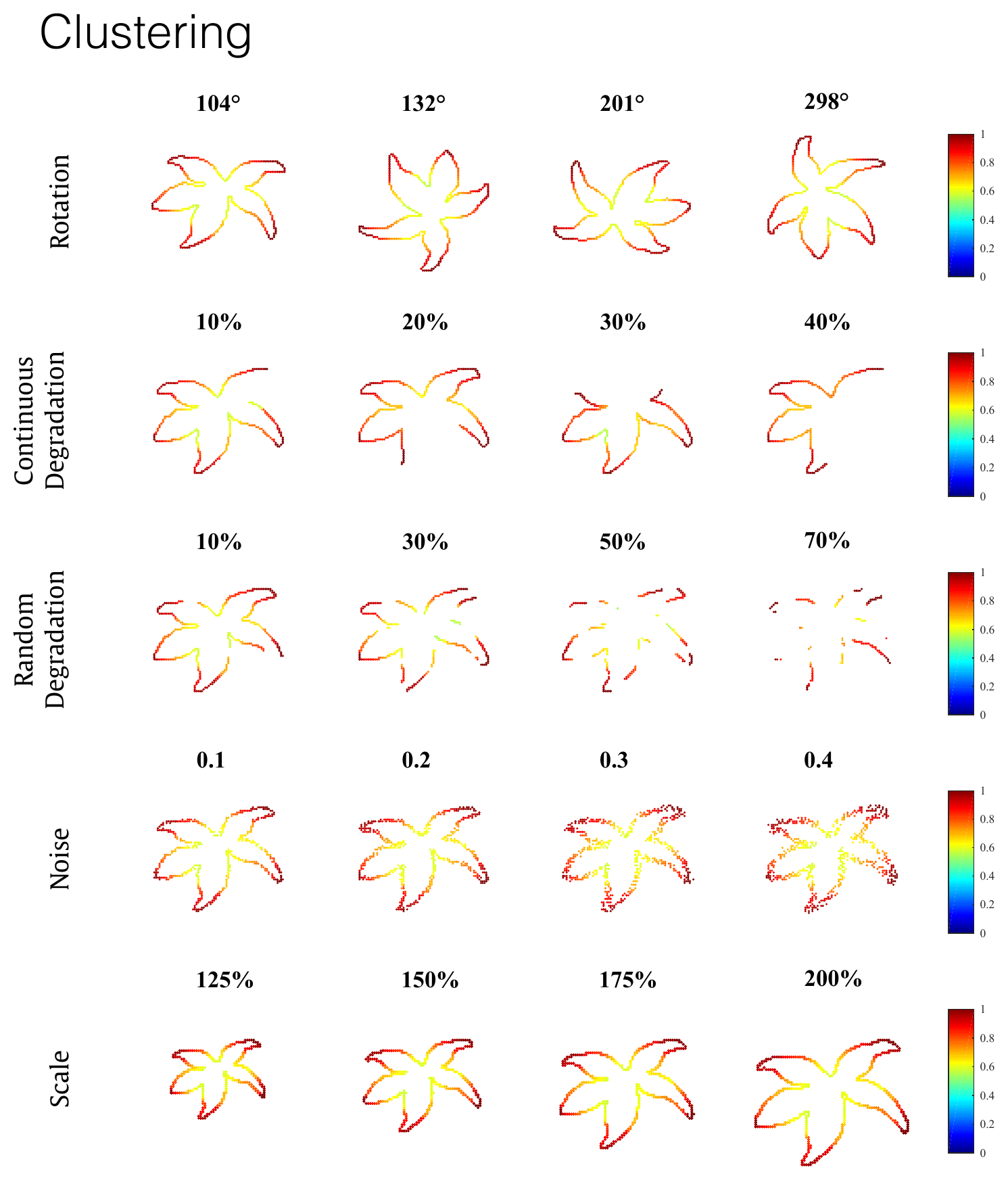}
\caption{Color-map representing the clustering coefficient ($cc$) of the contour pixels using threshold $T_l=0.325$ for the network construction. Each row illustrates a transformation and its respective parameters.}
\label{fig:clustHeatmap}
\end{figure}

\newpage
\begin{figure}[!ht]
\centering\includegraphics[scale=0.9]{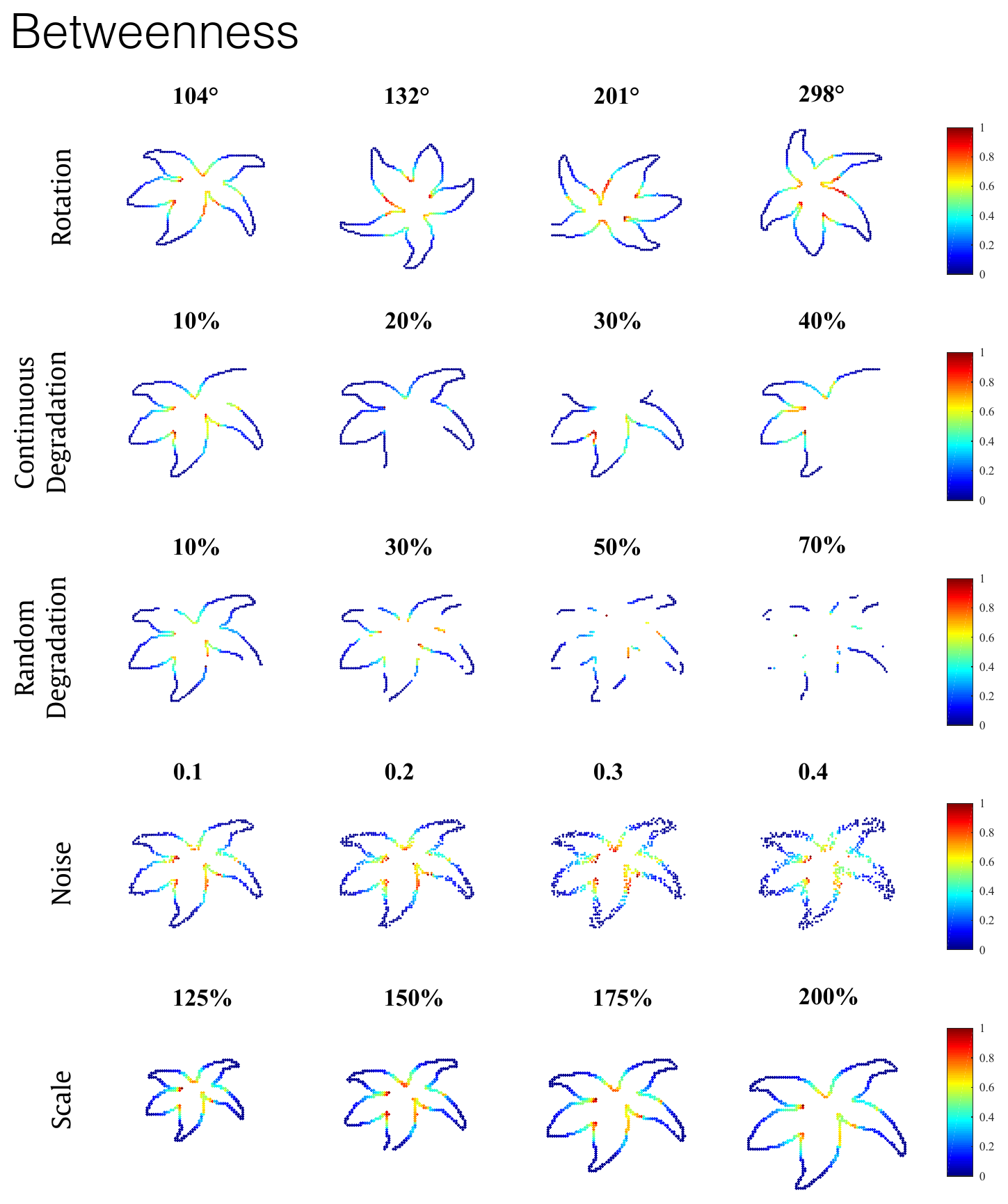}
\caption{Color-map representing the betweenness centrality ($b$) of the contour pixels using threshold $T_l=0.325$ for the network construction. Each row illustrates a transformation and its respective parameters.}
\label{fig:betwHeatmap}
\end{figure}

Figs.~\ref{fig:degreeHeatmap}, \ref{fig:clustHeatmap} and \ref{fig:betwHeatmap} present the distribution per network node (per pixel) of the following measurements: degree ($k$), clustering coefficient ($cc$) and betweenness ($b$). In this example, we used the $\delta_{T_l}(W)$ transformation, therefore, two pixels are connected to each other if the euclidean distance between them is \textit{smaller than} $T_l = 0.325$. The five transformations which were applied to the \textit{leaves} dataset are illustrated in each row. For what concerns the degree (Fig.~\ref{fig:degreeHeatmap}), we can see that the pixels that are closer to the leaf center are the most connected ones. In the case of the degraded contours, there are missing parts in the leaf folds, and, consequently, the degree of the pixels near those parts are smaller.

For the clustering coefficient (Fig.~\ref{fig:clustHeatmap}), we can observe that it is higher near the leaf tips. This can be explained by the fact that, at this threshold ($T_l = 0.325$), there are many connected triples among the pixels of the tips, which increases the $cc$ value. For the pixels near the leaf center there is also a high number of connected triples, however, their degree is also higher, which increases the number of possible connections among the neighbors of a pixel, but they are not always connected as triples. Consequently, the $cc$ value for those pixels is lower. Even for the degraded contours, a similar distribution pattern of the clustering coefficient can be observed. Yet, it should be noticed that for all the examples in Fig.~\ref{fig:clustHeatmap}, the clustering coefficient is already high, above 0.5.

Similarly to the degree color-map, we can observe that the betweenness centrality (Fig.~\ref{fig:betwHeatmap}) is also higher for the pixels near the leaf center, which are between the leaf tips. The pixels in that region belong to many pathways of the contour network, therefore, they connect pixels of different leaf tips. Consequently, their betweenness is also higher.   

%\newpage

In addition, we also performed an experiment to test the classification accuracy for what concerns the different variations of the \textit{leaves} dataset. In this experiment, each transformation (rotation, scale, noise and the two degradation types) was evaluated separately. The classification accuracy was calculated taking into account the 30 classes of the \textit{leaves} dataset. The threshold interval was sampled in 13 equally spaced values of $T_l$. Table~\ref{tab:leaves-accuracy} presents the results.

% \usepackage{multirow}
\begin{table}[!ht]
\centering
\footnotesize
\caption{Classification accuracy (\%) for the datasets \textit{leaves-rotated}, \textit{leaves-scaled} and \textit{leaves-noise}.}
\label{tab:leaves-accuracy}
\begin{tabular}{l|c|ccc}
\hline
\multirow{6}{*}{Scale} & Scale Factor & k-NN & SMO & NB \\ \cline{2-5} 
 & 1.25 & $86.55 \pm 0.51$ & $82.52 \pm 0.52$ & $73.64 \pm 0.54$ \\
 & 1.50 & $87.10 \pm 0.44$ & $83.44 \pm 0.56$ & $74.50 \pm 0.54$ \\
 & 1.75 & $86.93 \pm 0.49$ & $83.37 \pm 0.61$ & $75.18 \pm 0.46$ \\
 & 2.00 & $87.67 \pm 0.53$ & $83.92 \pm 0.58$ & $74.64 \pm 0.54$ \\ \hline
 %& All & $89.50 \pm 0.26$ & $87.19 \pm 0.16$ & $71.56 \pm 0.19$ \\ \hline
\multirow{8}{*}{Rotation} & Degree & k-NN & SMO & NB \\ \cline{2-5} 
 & 7$^{\circ}$ & $82.84 \pm 0.61$ & $80.65 \pm 0.58$ & $72.06 \pm 0.41$ \\
 & 35$^{\circ}$ & $82.06 \pm 0.60$ & $77.90 \pm 0.53$ & $72.07 \pm 0.48$ \\
 & 104$^{\circ}$ & $80.40 \pm 0.56$ & $79.12 \pm 0.66$ & $72.58 \pm 0.55$ \\
 & 132$^{\circ}$ & $85.37 \pm 0.53$ & $78.59 \pm 0.61$ & $72.25 \pm 0.48$ \\
 & 201$^{\circ}$ & $81.21 \pm 0.54$ & $76.80 \pm 0.55$ & $71.96 \pm 0.51$ \\
 & 298$^{\circ}$ & $81.85 \pm 0.59$ & $77.04 \pm 0.59$ & $72.28 \pm 0.49$ \\ \hline
 %& All & $97.52 \pm 0.14$ & $87.86 \pm 0.15$ & $72.98 \pm 0.14$ \\ \hline
\multirow{6}{*}{Noise} & Noise level & k-NN & SMO & NB \\ \cline{2-5} 
 & 1 & $77.08 \pm 0.59$ & $76.25 \pm 0.68$ & $68.30 \pm 0.56$ \\
 & 2 & $75.82 \pm 0.68$ & $78.43 \pm 0.47$ & $68.35 \pm 0.56$ \\
 & 3 & $75.42 \pm 0.59$ & $77.36 \pm 0.59$ & $67.00 \pm 0.67$ \\
 & 4 & $76.54 \pm 0.58$ & $76.84 \pm 0.64$ & $67.87 \pm 0.61$ \\ \hline
 \multirow{2}{*}{Original} &  & k-NN & SMO & NB \\ \cline{2-5} 
 &  & $84.81 \pm 0.51$ & $79.99 \pm 0.62$ & $72.22 \pm 0.52$ \\ \hline
 %& All & $86.48 \pm 0.26$ & $82.72 \pm 0.23$ & $67.83 \pm 0.25$ \\ \hline
\end{tabular}
\end{table}

For what concerns rotation, the resulting network for a given threshold $T_l$ will be the same before and after the application of the transformation. The rotation does not affect the adjacency matrix of the network, and, consequently, the descriptor will be preserved. The differences observed for the accuracy values given the different rotation degrees (Table~\ref{tab:leaves-accuracy}) are due to the random sampling of the folds during the cross-validation. The descriptor is also not affected by linear transformations such as scale. Therefore, we can observe in Table~\ref{tab:leaves-accuracy} that the performance for rotation and scale transformations are similar to the performance obtained for the original dataset (\textit{leaves}). In the case of the scale transformation, the accuracy improved for all the classifiers, using the same feature vector $\Phi$. Therefore, the interpolation process associated with the scale transformation provided a better separation of the classes for the \textit{leaves} dataset. Yet, the measurements extracted from the network topology were more discriminative after the application of the scale transformation. Regarding the robustness to noise, we can see that the accuracy for the \textit{leaves-noise} dataset is lower when compared with the other two datasets presented in the this table. The decrease in accuracy was the highest when compared to the other datasets. However, the accuracy is not much affected by the different levels of noise.
%In addition, changing the measurements extracted from the network topology did not influence the robustness to noise property of the descriptor, which means that this property is an intrinsic characteristic of the network representation of the shape boundary. Finally, it is expected a decrease in classification accuracy as the degradation levels increase because the shape boundaries 

%\textcolor{red}{Comentar porque a acuracia melhorou para a rotacao e a escala, e tambem, porque que quando o ruido aumenta a acuracia tambem melhora... talvez porque as transformacoes aplicadas nas imagens resultaram de alguma forma numa maior separacao entre as classes dado o mesmo conjunto de medidas extraidas das redes. De alguma forma as medidas foram mais discriminativas para o cojunto das folhas depois que as transformacoes foram aplicadas.}

\begin{figure}[!h]
\centering\includegraphics[scale=0.5]{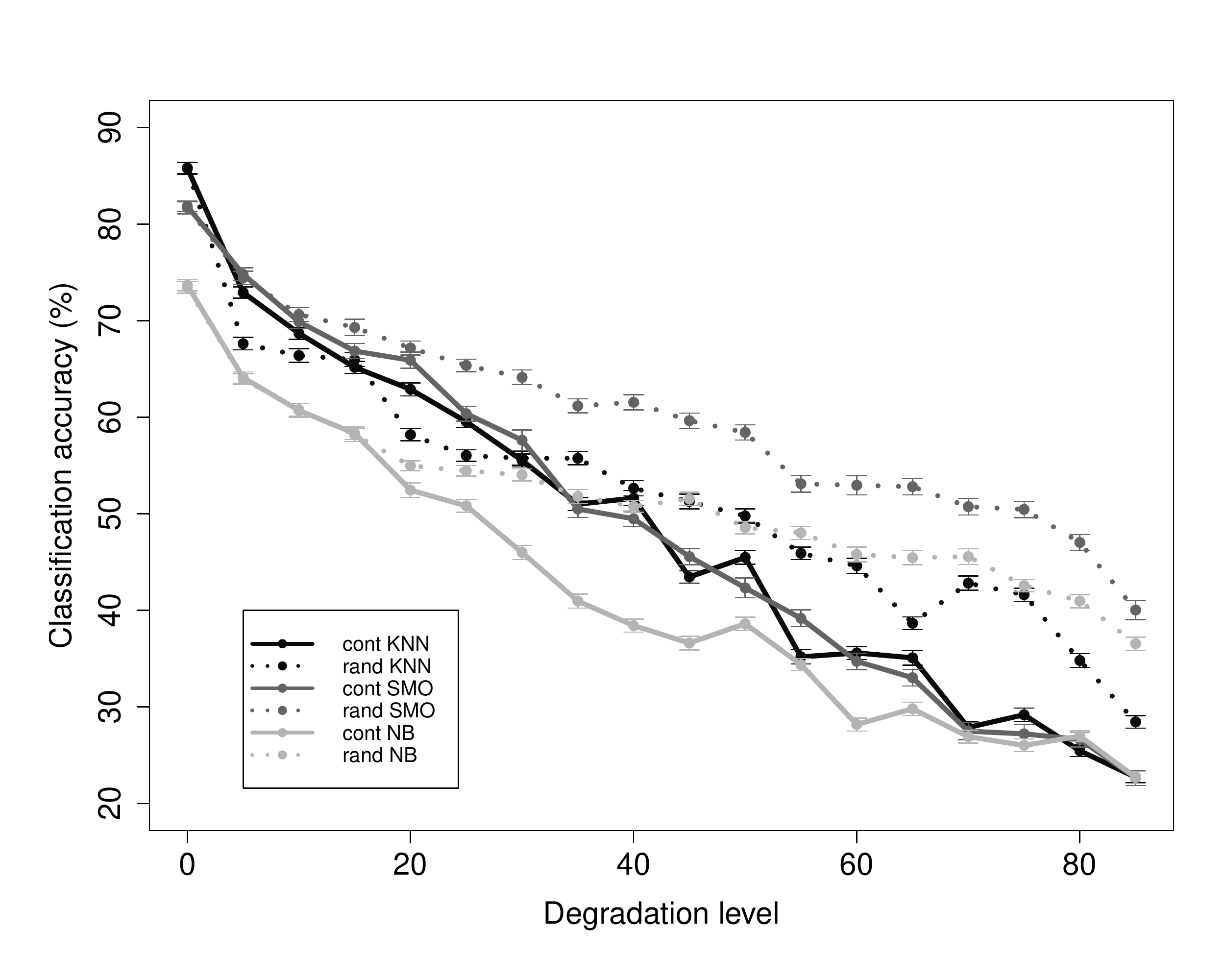}
\caption{Classification accuracy as a function of the degradation levels for the \textit{leaves-degraded} dataset.}
\label{fig:degradationPlot}
\end{figure}

Fig~\ref{fig:degradationPlot} presents the classification accuracy (\%) as a function of the different degradation levels for the \textit{leaves-degraded} dataset. As the degradation level increases, the classification performance decreases. As expected, when parts of the shape boundaries are removed, the correct identification of the classes becomes more difficult.
%, which is corroborated by the plots of F

%\subsection{\textcolor{red}{Comparison with other shape classification methods}}

\section{Conclusion}
\label{sec:conclusion}
 
We presented in this paper an approach for shape analysis aiming at pattern recognition applications. This approach extends the method of Backes \textit{et al.}~\cite{backes2009complex} in what concerns the structural analysis performed over the network obtained through shape boundaries. Therefore, we went a step further performing a study of the relationship between the shape architecture and the network topology. This was first investigated for regular polygons and then for other irregular and generic shapes. Through the use of the curvature analysis, it was possible to investigate how the network measurements vary according to some specific shape properties, and, consequently, how the network descriptor, $\Phi$ is constructed. In addition, we demonstrated the effect of the threshold sampling in the classification accuracy regarding the dynamical construction of $\Phi$ (Fig.~\ref{fig:accuracy}), and, also, the accuracy provided by each threshold individually (Fig.~\ref{fig:singleThreshold}).

Through the use of a more robust set of structural measurements we could improve the classification performance in different pattern recognition problems. The \textit{geom-shapes} and the \textit{generic-shapes} datasets represent applications that can be used as a benchmark comparison. We have shown that, for both datasets, the proposed shape descriptor can improve the classification performance when compared to the CNDescriptor for the different comparisons considered in the experiments (\textit{smaller-than} and \textit{greater-than}), and, also for different numbers of thresholds and features. The increase observed in the classification accuracy also depends on the classifier. The SMO classifier provided the highest improvement regarding the classification performance (\%). In spite of that, the NB classifier achieved a slightly better performance for those datasets when using CNDescriptor. For what concerns the \textit{leaves} dataset, there was a considerable increase in the classification accuracy for all classifiers and for both comparisons when using the proposed feature vector, $\Phi$. It should also be noticed that when using a smaller number of descriptors for this dataset the increase in classification performance is even higher for $\Phi$ than for the CNDescriptor. We consider \textit{leaves} a more challenging dataset both because the number of classes (30) is higher than the number of samples per class (20), and, also because because of the intrinsic nature of the samples, which are part of a real-world application. In addition, Backes \textit{et al.}~\cite{backes2009complex} performed a comparison of the accuracy obtained with CNDescriptor with classical methods found in literature for shape analysis, e.g., Fourier descriptors, Zernike moments, curvature descriptors and others. They showed that their method could surpass these classical approaches for what concerns the classification accuracy (\%). In this paper we have shown that a more robust set of network measurements could even surpass the CNDescriptor for the \textit{leaves} dataset as well as in many comparisons performed for the other two datasets.

Finally, one important contribution of this paper is related to the construction of the feature vector from the network obtained from shape boundaries. We have shown that not only degree related measurements are capable of distinguishing categories associated with a specific problem. In addition, we also demonstrated how other classes of measurements can be used together in pattern recognition applications. Yet, when using a set of measurements that accounts for different structural properties of networks, the number of thresholds can be reduced, and, consequently, the construction of the shape descriptor becomes more computationally efficient. Therefore, there is a balance between the amount of measurements and thresholds that should be selected to compose the feature vector.

%\textcolor{red}{Na base das folhas, por exemplo, fica evidente que o nosso metodo eh melhor quando eh utilizado um numero menor de thresholds, e, usar um numero menor de atributos eh computacionalmente mais rápido.... na verdade acho que isso pode ser discutido na conclusao, seria algo do tipo a necessidade de um balanço entre as medidas escolhidas para se compor o vetor de atributos e o numero de thresholds... quero dizer, se escolhermos boas medidas talvez nao precise de tantos thresholds, alem de demorar menos pra rodar}

%\textcolor{red}{Falar que diferentemente do trabalho do Backes, além de generalizar o vetor de caracteristicas, nos testamos outros coisas do método tais como a acurácia para diferentes amostragens do intevalo de threhsholds, ou seja, o efeito da amostragem do intervalo na acuracia final; o threshold unico; diferentes tipos de classificadores (explorar mais isso) 
%\textcolor{red}{Falar tambem que no trabalho deles ja foi feita a comparacao com outros metodos classicos da literatura e por isso neste trabalho fizemos uma comparacao exaustiva do nosso metodo com o metodo deles?}

\section*{Acknowledgements}

G.H.B.M. and O.M.B. are grateful for the support from S\~{a}o Paulo Research Foundation (FAPESP) with grant \#2015/05899-7. G.H.B.M. and J.M. are grateful for the support of the Coordination for the Improvement of Higher Education Personnel (CAPES). O.M.B. gratefully acknowledges the financial support of National Council for Scientific and Technological Development (CNPq), grants \#307797/2014-7 and \#484312/2013-8, and, FAPESP with grant \#2014/08026-1.

% \begin{table}[h]
% \centering
% \begin{tabular}{l l l}
% \hline
% \textbf{Treatments} & \textbf{Response 1} & \textbf{Response 2}\\
% \hline
% Treatment 1 & 0.0003262 & 0.562 \\
% Treatment 2 & 0.0015681 & 0.910 \\
% Treatment 3 & 0.0009271 & 0.296 \\
% \hline
% \end{tabular}
% \caption{Table caption}
% \end{table}

%\subsection{Subsection Two}
% \begin{figure}[h]
% \centering\includegraphics[width=0.4\linewidth]{placeholder}
% \caption{Figure caption}
% \end{figure}


\section*{References}

%% References


%\bibliography{sample.bib}

\begin{thebibliography}{46}
\expandafter\ifx\csname natexlab\endcsname\relax\def\natexlab#1{#1}\fi
\providecommand{\bibinfo}[2]{#2}
\ifx\xfnm\relax \def\xfnm[#1]{\unskip,\space#1}\fi
%Type = Inproceedings
\bibitem[{Gerig et~al.(2001)Gerig, Styner, Shenton, and
  Lieberman}]{gerig2001shape}
\bibinfo{author}{G.~Gerig}, \bibinfo{author}{M.~Styner}, \bibinfo{author}{M.~E.
  Shenton}, \bibinfo{author}{J.~A. Lieberman},
\newblock \bibinfo{title}{Shape versus size: Improved understanding of the
  morphology of brain structures},
\newblock in: \bibinfo{booktitle}{International Conference on Medical Image
  Computing and Computer-Assisted Intervention},
  \bibinfo{organization}{Springer}, pp. \bibinfo{pages}{24--32}.
%Type = Article
\bibitem[{Torres et~al.(2004)Torres, Falcao, and Costa}]{torres2004graph}
\bibinfo{author}{R.~d.~S. Torres}, \bibinfo{author}{A.~X. Falcao},
  \bibinfo{author}{L.~d.~F. Costa},
\newblock \bibinfo{title}{A graph-based approach for multiscale shape
  analysis},
\newblock \bibinfo{journal}{Pattern Recognition} \bibinfo{volume}{37}
  (\bibinfo{year}{2004}) \bibinfo{pages}{1163--1174}.
%Type = Article
\bibitem[{Fan et~al.(2007)Fan, Shen, Gur, Gur, and Davatzikos}]{fan2007compare}
\bibinfo{author}{Y.~Fan}, \bibinfo{author}{D.~Shen}, \bibinfo{author}{R.~C.
  Gur}, \bibinfo{author}{R.~E. Gur}, \bibinfo{author}{C.~Davatzikos},
\newblock \bibinfo{title}{Compare: classification of morphological patterns
  using adaptive regional elements},
\newblock \bibinfo{journal}{IEEE transactions on medical imaging}
  \bibinfo{volume}{26} (\bibinfo{year}{2007}) \bibinfo{pages}{93--105}.
%Type = Article
\bibitem[{Aitkenhead et~al.(2003)Aitkenhead, Dalgetty, Mullins, McDonald, and
  Strachan}]{aitkenhead2003weed}
\bibinfo{author}{M.~Aitkenhead}, \bibinfo{author}{I.~Dalgetty},
  \bibinfo{author}{C.~Mullins}, \bibinfo{author}{A.~J.~S. McDonald},
  \bibinfo{author}{N.~J.~C. Strachan},
\newblock \bibinfo{title}{Weed and crop discrimination using image analysis and
  artificial intelligence methods},
\newblock \bibinfo{journal}{Computers and electronics in Agriculture}
  \bibinfo{volume}{39} (\bibinfo{year}{2003}) \bibinfo{pages}{157--171}.
%Type = Article
\bibitem[{Neto et~al.(2006)Neto, Meyer, Jones, and Samal}]{neto2006plant}
\bibinfo{author}{J.~C. Neto}, \bibinfo{author}{G.~E. Meyer},
  \bibinfo{author}{D.~D. Jones}, \bibinfo{author}{A.~K. Samal},
\newblock \bibinfo{title}{Plant species identification using elliptic fourier
  leaf shape analysis},
\newblock \bibinfo{journal}{Computers and electronics in agriculture}
  \bibinfo{volume}{50} (\bibinfo{year}{2006}) \bibinfo{pages}{121--134}.
%Type = Article
\bibitem[{Brosnan and Sun(2002)}]{brosnan2002inspection}
\bibinfo{author}{T.~Brosnan}, \bibinfo{author}{D.-W. Sun},
\newblock \bibinfo{title}{Inspection and grading of agricultural and food
  products by computer vision systems—a review},
\newblock \bibinfo{journal}{Computers and electronics in agriculture}
  \bibinfo{volume}{36} (\bibinfo{year}{2002}) \bibinfo{pages}{193--213}.
%Type = Article
\bibitem[{Costa et~al.(2011)Costa, Antonucci, Pallottino, Aguzzi, Sun, and
  Menesatti}]{costa2011shape}
\bibinfo{author}{C.~Costa}, \bibinfo{author}{F.~Antonucci},
  \bibinfo{author}{F.~Pallottino}, \bibinfo{author}{J.~Aguzzi},
  \bibinfo{author}{D.-W. Sun}, \bibinfo{author}{P.~Menesatti},
\newblock \bibinfo{title}{Shape analysis of agricultural products: a review of
  recent research advances and potential application to computer vision},
\newblock \bibinfo{journal}{Food and Bioprocess Technology} \bibinfo{volume}{4}
  (\bibinfo{year}{2011}) \bibinfo{pages}{673--692}.
%Type = Article
\bibitem[{Hemming and Rath(2001)}]{hemming2001pa}
\bibinfo{author}{J.~Hemming}, \bibinfo{author}{T.~Rath},
\newblock \bibinfo{title}{Pa—precision agriculture: Computer-vision-based
  weed identification under field conditions using controlled lighting},
\newblock \bibinfo{journal}{Journal of agricultural engineering research}
  \bibinfo{volume}{78} (\bibinfo{year}{2001}) \bibinfo{pages}{233--243}.
%Type = Article
\bibitem[{Plotze et~al.(2005)Plotze, Falvo, P{\'a}dua, Bernacci, Vieira,
  Oliveira, and Bruno}]{plotze2005leaf}
\bibinfo{author}{R.~d.~O. Plotze}, \bibinfo{author}{M.~Falvo},
  \bibinfo{author}{J.~G. P{\'a}dua}, \bibinfo{author}{L.~C. Bernacci},
  \bibinfo{author}{M.~L.~C. Vieira}, \bibinfo{author}{G.~C.~X. Oliveira},
  \bibinfo{author}{O.~M. Bruno},
\newblock \bibinfo{title}{Leaf shape analysis using the multiscale minkowski
  fractal dimension, a new morphometric method: a study with passiflora
  (passifloraceae)},
\newblock \bibinfo{journal}{Canadian Journal of Botany} \bibinfo{volume}{83}
  (\bibinfo{year}{2005}) \bibinfo{pages}{287--301}.
%Type = Article
\bibitem[{Rangayyan et~al.(2000)Rangayyan, Mudigonda, and
  Desautels}]{rangayyan2000boundary}
\bibinfo{author}{R.~M. Rangayyan}, \bibinfo{author}{N.~R. Mudigonda},
  \bibinfo{author}{J.~L. Desautels},
\newblock \bibinfo{title}{Boundary modelling and shape analysis methods for
  classification of mammographic masses},
\newblock \bibinfo{journal}{Medical and Biological Engineering and Computing}
  \bibinfo{volume}{38} (\bibinfo{year}{2000}) \bibinfo{pages}{487--496}.
%Type = Article
\bibitem[{Heimann and Meinzer(2009)}]{heimann2009statistical}
\bibinfo{author}{T.~Heimann}, \bibinfo{author}{H.-P. Meinzer},
\newblock \bibinfo{title}{Statistical shape models for 3d medical image
  segmentation: a review},
\newblock \bibinfo{journal}{Medical image analysis} \bibinfo{volume}{13}
  (\bibinfo{year}{2009}) \bibinfo{pages}{543--563}.
%Type = Article
\bibitem[{Selle et~al.(2002)Selle, Preim, Schenk, and
  Peitgen}]{selle2002analysis}
\bibinfo{author}{D.~Selle}, \bibinfo{author}{B.~Preim},
  \bibinfo{author}{A.~Schenk}, \bibinfo{author}{H.-O. Peitgen},
\newblock \bibinfo{title}{Analysis of vasculature for liver surgical planning},
\newblock \bibinfo{journal}{IEEE transactions on medical imaging}
  \bibinfo{volume}{21} (\bibinfo{year}{2002}) \bibinfo{pages}{1344--1357}.
%Type = Article
\bibitem[{Tsai et~al.(2003)Tsai, Yezzi, Wells, Tempany, Tucker, Fan, Grimson,
  and Willsky}]{tsai2003shape}
\bibinfo{author}{A.~Tsai}, \bibinfo{author}{A.~Yezzi},
  \bibinfo{author}{W.~Wells}, \bibinfo{author}{C.~Tempany},
  \bibinfo{author}{D.~Tucker}, \bibinfo{author}{A.~Fan}, \bibinfo{author}{W.~E.
  Grimson}, \bibinfo{author}{A.~Willsky},
\newblock \bibinfo{title}{A shape-based approach to the segmentation of medical
  imagery using level sets},
\newblock \bibinfo{journal}{IEEE transactions on medical imaging}
  \bibinfo{volume}{22} (\bibinfo{year}{2003}) \bibinfo{pages}{137--154}.
%Type = Article
\bibitem[{Benz et~al.(2004)Benz, Hofmann, Willhauck, Lingenfelder, and
  Heynen}]{benz2004multi}
\bibinfo{author}{U.~C. Benz}, \bibinfo{author}{P.~Hofmann},
  \bibinfo{author}{G.~Willhauck}, \bibinfo{author}{I.~Lingenfelder},
  \bibinfo{author}{M.~Heynen},
\newblock \bibinfo{title}{Multi-resolution, object-oriented fuzzy analysis of
  remote sensing data for gis-ready information},
\newblock \bibinfo{journal}{ISPRS Journal of photogrammetry and remote sensing}
  \bibinfo{volume}{58} (\bibinfo{year}{2004}) \bibinfo{pages}{239--258}.
%Type = Article
\bibitem[{Blaschke(2010)}]{blaschke2010object}
\bibinfo{author}{T.~Blaschke},
\newblock \bibinfo{title}{Object based image analysis for remote sensing},
\newblock \bibinfo{journal}{ISPRS journal of photogrammetry and remote sensing}
  \bibinfo{volume}{65} (\bibinfo{year}{2010}) \bibinfo{pages}{2--16}.
%Type = Article
\bibitem[{Osowski et~al.(2002)}]{osowski2002fourier}
\bibinfo{author}{S.~Osowski}, et~al.,
\newblock \bibinfo{title}{Fourier and wavelet descriptors for shape recognition
  using neural networks—a comparative study},
\newblock \bibinfo{journal}{Pattern Recognition} \bibinfo{volume}{35}
  (\bibinfo{year}{2002}) \bibinfo{pages}{1949--1957}.
%Type = Inproceedings
\bibitem[{Zhang et~al.(2001)Zhang, Lu et~al.}]{zhang2001comparative}
\bibinfo{author}{D.~Zhang}, \bibinfo{author}{G.~Lu}, et~al.,
\newblock \bibinfo{title}{A comparative study on shape retrieval using fourier
  descriptors with different shape signatures},
\newblock in: \bibinfo{booktitle}{Proc. of international conference on
  intelligent multimedia and distance education (ICIMADE01)}, pp.
  \bibinfo{pages}{1--9}.
%Type = Article
\bibitem[{Davatzikos et~al.(2003)Davatzikos, Tao, and
  Shen}]{davatzikos2003hierarchical}
\bibinfo{author}{C.~Davatzikos}, \bibinfo{author}{X.~Tao},
  \bibinfo{author}{D.~Shen},
\newblock \bibinfo{title}{Hierarchical active shape models, using the wavelet
  transform},
\newblock \bibinfo{journal}{IEEE transactions on medical imaging}
  \bibinfo{volume}{22} (\bibinfo{year}{2003}) \bibinfo{pages}{414--423}.
%Type = Article
\bibitem[{Soares et~al.(2006)Soares, Leandro, Cesar, Jelinek, and
  Cree}]{soares2006retinal}
\bibinfo{author}{J.~V. Soares}, \bibinfo{author}{J.~J. Leandro},
  \bibinfo{author}{R.~M. Cesar}, \bibinfo{author}{H.~F. Jelinek},
  \bibinfo{author}{M.~J. Cree},
\newblock \bibinfo{title}{Retinal vessel segmentation using the 2-d gabor
  wavelet and supervised classification},
\newblock \bibinfo{journal}{IEEE Transactions on medical Imaging}
  \bibinfo{volume}{25} (\bibinfo{year}{2006}) \bibinfo{pages}{1214--1222}.
%Type = Article
\bibitem[{Backes and Bruno(2010)}]{backes2010shape}
\bibinfo{author}{A.~R. Backes}, \bibinfo{author}{O.~M. Bruno},
\newblock \bibinfo{title}{Shape classification using complex network and
  multi-scale fractal dimension},
\newblock \bibinfo{journal}{Pattern Recognition Letters} \bibinfo{volume}{31}
  (\bibinfo{year}{2010}) \bibinfo{pages}{44--51}.
%Type = Article
\bibitem[{Backes et~al.(2012)Backes, Florindo, and Bruno}]{backes2012shape}
\bibinfo{author}{A.~R. Backes}, \bibinfo{author}{J.~B. Florindo},
  \bibinfo{author}{O.~M. Bruno},
\newblock \bibinfo{title}{Shape analysis using fractal dimension: A curvature
  based approach},
\newblock \bibinfo{journal}{Chaos: An Interdisciplinary Journal of Nonlinear
  Science} \bibinfo{volume}{22} (\bibinfo{year}{2012}) \bibinfo{pages}{043103}.
%Type = Article
\bibitem[{Bruno et~al.(2008)Bruno, de~Oliveira~Plotze, Falvo, and
  de~Castro}]{bruno2008fractal}
\bibinfo{author}{O.~M. Bruno}, \bibinfo{author}{R.~de~Oliveira~Plotze},
  \bibinfo{author}{M.~Falvo}, \bibinfo{author}{M.~de~Castro},
\newblock \bibinfo{title}{Fractal dimension applied to plant identification},
\newblock \bibinfo{journal}{Information Sciences} \bibinfo{volume}{178}
  (\bibinfo{year}{2008}) \bibinfo{pages}{2722--2733}.
%Type = Book
\bibitem[{Mokhtarian and Bober(2013)}]{mokhtarian2013curvature}
\bibinfo{author}{F.~Mokhtarian}, \bibinfo{author}{M.~Bober},
  \bibinfo{title}{Curvature scale space representation: theory, applications,
  and MPEG-7 standardization}, volume~\bibinfo{volume}{25},
  \bibinfo{publisher}{Springer Science \& Business Media},
  \bibinfo{year}{2013}.
%Type = Article
\bibitem[{Backes et~al.(2009)Backes, Casanova, and Bruno}]{backes2009complex}
\bibinfo{author}{A.~R. Backes}, \bibinfo{author}{D.~Casanova},
  \bibinfo{author}{O.~M. Bruno},
\newblock \bibinfo{title}{A complex network-based approach for boundary shape
  analysis},
\newblock \bibinfo{journal}{Pattern Recognition} \bibinfo{volume}{42}
  (\bibinfo{year}{2009}) \bibinfo{pages}{54--67}.
%Type = Article
\bibitem[{Backes and Bruno(2013)}]{backes2013polygonal}
\bibinfo{author}{A.~R. Backes}, \bibinfo{author}{O.~M. Bruno},
\newblock \bibinfo{title}{Polygonal approximation of digital planar curves
  through vertex betweenness},
\newblock \bibinfo{journal}{Information Sciences} \bibinfo{volume}{222}
  (\bibinfo{year}{2013}) \bibinfo{pages}{795--804}.
%Type = Article
\bibitem[{Wu et~al.(2015)Wu, Lu, and Deng}]{wu2015image}
\bibinfo{author}{Z.~Wu}, \bibinfo{author}{X.~Lu}, \bibinfo{author}{Y.~Deng},
\newblock \bibinfo{title}{Image edge detection based on local dimension: A
  complex networks approach},
\newblock \bibinfo{journal}{Physica A: Statistical Mechanics and its
  Applications} \bibinfo{volume}{440} (\bibinfo{year}{2015})
  \bibinfo{pages}{9--18}.
%Type = Article
\bibitem[{Sui et~al.(2016)Sui, Shao, Wang, Sun, and Ji}]{sui2016complex}
\bibinfo{author}{Y.~Sui}, \bibinfo{author}{F.~Shao}, \bibinfo{author}{C.~Wang},
  \bibinfo{author}{R.~Sun}, \bibinfo{author}{J.~Ji},
\newblock \bibinfo{title}{Complex network modeling of spectral remotely sensed
  imagery: A case study of massive green algae blooms detection based on modis
  data},
\newblock \bibinfo{journal}{Physica A: Statistical Mechanics and its
  Applications} \bibinfo{volume}{464} (\bibinfo{year}{2016})
  \bibinfo{pages}{138--148}.
%Type = Article
\bibitem[{Backes et~al.(2013)Backes, Casanova, and Bruno}]{backes2013texture}
\bibinfo{author}{A.~R. Backes}, \bibinfo{author}{D.~Casanova},
  \bibinfo{author}{O.~M. Bruno},
\newblock \bibinfo{title}{Texture analysis and classification: A complex
  network-based approach},
\newblock \bibinfo{journal}{Information Sciences} \bibinfo{volume}{219}
  (\bibinfo{year}{2013}) \bibinfo{pages}{168--180}.
%Type = Article
\bibitem[{Newman(2003)}]{newman2003structure}
\bibinfo{author}{M.~E. Newman},
\newblock \bibinfo{title}{The structure and function of complex networks},
\newblock \bibinfo{journal}{SIAM review} \bibinfo{volume}{45}
  (\bibinfo{year}{2003}) \bibinfo{pages}{167--256}.
%Type = Article
\bibitem[{Boccaletti et~al.(2006)Boccaletti, Latora, Moreno, Chavez, and
  Hwang}]{boccaletti2006complex}
\bibinfo{author}{S.~Boccaletti}, \bibinfo{author}{V.~Latora},
  \bibinfo{author}{Y.~Moreno}, \bibinfo{author}{M.~Chavez},
  \bibinfo{author}{D.-U. Hwang},
\newblock \bibinfo{title}{Complex networks: Structure and dynamics},
\newblock \bibinfo{journal}{Physics reports} \bibinfo{volume}{424}
  (\bibinfo{year}{2006}) \bibinfo{pages}{175--308}.
%Type = Book
\bibitem[{Newman et~al.(2011)Newman, Barabasi, and Watts}]{newman2011structure}
\bibinfo{author}{M.~Newman}, \bibinfo{author}{A.-L. Barabasi},
  \bibinfo{author}{D.~J. Watts}, \bibinfo{title}{The structure and dynamics of
  networks}, \bibinfo{publisher}{Princeton University Press},
  \bibinfo{year}{2011}.
%Type = Article
\bibitem[{Barab{\'a}si and Albert(1999)}]{barabasi1999emergence}
\bibinfo{author}{A.-L. Barab{\'a}si}, \bibinfo{author}{R.~Albert},
\newblock \bibinfo{title}{Emergence of scaling in random networks},
\newblock \bibinfo{journal}{science} \bibinfo{volume}{286}
  (\bibinfo{year}{1999}) \bibinfo{pages}{509--512}.
%Type = Article
\bibitem[{Barab{\'a}si et~al.(2000)Barab{\'a}si, Albert, and
  Jeong}]{barabasi2000scale}
\bibinfo{author}{A.-L. Barab{\'a}si}, \bibinfo{author}{R.~Albert},
  \bibinfo{author}{H.~Jeong},
\newblock \bibinfo{title}{Scale-free characteristics of random networks: the
  topology of the world-wide web},
\newblock \bibinfo{journal}{Physica A: Statistical Mechanics and its
  Applications} \bibinfo{volume}{281} (\bibinfo{year}{2000})
  \bibinfo{pages}{69--77}.
%Type = Article
\bibitem[{Watts and Strogatz(1998)}]{watts1998collective}
\bibinfo{author}{D.~J. Watts}, \bibinfo{author}{S.~H. Strogatz},
\newblock \bibinfo{title}{Collective dynamics of 'small-world' networks},
\newblock \bibinfo{journal}{nature} \bibinfo{volume}{393}
  (\bibinfo{year}{1998}) \bibinfo{pages}{440--442}.
%Type = Article
\bibitem[{Costa et~al.(2007)Costa, Rodrigues, Travieso, and
  Villas~Boas}]{costa2007characterization}
\bibinfo{author}{L.~d.~F. Costa}, \bibinfo{author}{F.~A. Rodrigues},
  \bibinfo{author}{G.~Travieso}, \bibinfo{author}{P.~R. Villas~Boas},
\newblock \bibinfo{title}{Characterization of complex networks: A survey of
  measurements},
\newblock \bibinfo{journal}{Advances in physics} \bibinfo{volume}{56}
  (\bibinfo{year}{2007}) \bibinfo{pages}{167--242}.
%Type = Article
\bibitem[{Newman(2005)}]{newman2005measure}
\bibinfo{author}{M.~E. Newman},
\newblock \bibinfo{title}{A measure of betweenness centrality based on random
  walks},
\newblock \bibinfo{journal}{Social networks} \bibinfo{volume}{27}
  (\bibinfo{year}{2005}) \bibinfo{pages}{39--54}.
%Type = Article
\bibitem[{Borgatti(2005)}]{borgatti2005centrality}
\bibinfo{author}{S.~P. Borgatti},
\newblock \bibinfo{title}{Centrality and network flow},
\newblock \bibinfo{journal}{Social networks} \bibinfo{volume}{27}
  (\bibinfo{year}{2005}) \bibinfo{pages}{55--71}.
%Type = Article
\bibitem[{da~F~Costa et~al.(2010)da~F~Costa, Boas, Silva, and
  Rodrigues}]{da2010pattern}
\bibinfo{author}{L.~da~F~Costa}, \bibinfo{author}{P.~V. Boas},
  \bibinfo{author}{F.~Silva}, \bibinfo{author}{F.~Rodrigues},
\newblock \bibinfo{title}{A pattern recognition approach to complex networks},
\newblock \bibinfo{journal}{Journal of Statistical Mechanics: Theory and
  Experiment} \bibinfo{volume}{2010} (\bibinfo{year}{2010})
  \bibinfo{pages}{P11015}.
%Type = Inproceedings
\bibitem[{Sharvit et~al.(1998)Sharvit, Chan, Tek, and
  Kimia}]{sharvit1998symmetry}
\bibinfo{author}{D.~Sharvit}, \bibinfo{author}{J.~Chan},
  \bibinfo{author}{H.~Tek}, \bibinfo{author}{B.~B. Kimia},
\newblock \bibinfo{title}{Symmetry-based indexing of image databases},
\newblock in: \bibinfo{booktitle}{Content-Based Access of Image and Video
  Libraries, 1998. Proceedings. IEEE Workshop on},
  \bibinfo{organization}{IEEE}, pp. \bibinfo{pages}{56--62}.
%Type = Article
\bibitem[{Sebastian et~al.(2004)Sebastian, Klein, and
  Kimia}]{sebastian2004recognition}
\bibinfo{author}{T.~B. Sebastian}, \bibinfo{author}{P.~N. Klein},
  \bibinfo{author}{B.~B. Kimia},
\newblock \bibinfo{title}{Recognition of shapes by editing their shock graphs},
\newblock \bibinfo{journal}{IEEE Transactions on pattern analysis and machine
  intelligence} \bibinfo{volume}{26} (\bibinfo{year}{2004})
  \bibinfo{pages}{550--571}.
%Type = Article
\bibitem[{Newman(2002)}]{newman2002assortative}
\bibinfo{author}{M.~E. Newman},
\newblock \bibinfo{title}{Assortative mixing in networks},
\newblock \bibinfo{journal}{Physical review letters} \bibinfo{volume}{89}
  (\bibinfo{year}{2002}) \bibinfo{pages}{208701}.
%Type = Article
\bibitem[{Attneave(1954)}]{attneave1954some}
\bibinfo{author}{F.~Attneave},
\newblock \bibinfo{title}{Some informational aspects of visual perception.},
\newblock \bibinfo{journal}{Psychological review} \bibinfo{volume}{61}
  (\bibinfo{year}{1954}) \bibinfo{pages}{183}.
%Type = Article
\bibitem[{Cesar and Costa(1996)}]{cesar1996towards}
\bibinfo{author}{R.~M. Cesar}, \bibinfo{author}{L.~D.~F. Costa},
\newblock \bibinfo{title}{Towards effective planar shape representation with
  multiscale digital curvature analysis based on signal processing techniques},
\newblock \bibinfo{journal}{Pattern Recognition} \bibinfo{volume}{29}
  (\bibinfo{year}{1996}) \bibinfo{pages}{1559--1569}.
%Type = Misc
\bibitem[{lea(2009)}]{leavesDataset}
\bibinfo{title}{{Leaves dataset}},
  \bibinfo{howpublished}{\url{http://scg.ifsc.usp.br/dataset/ShapeCN.php}},
  \bibinfo{year}{2009}. \bibinfo{note}{[Online; accessed 30-October-2017]}.
%Type = Incollection
\bibitem[{Platt(1998)}]{Platt1998}
\bibinfo{author}{J.~Platt},
\newblock \bibinfo{title}{Fast training of support vector machines using
  sequential minimal optimization},
\newblock in: \bibinfo{editor}{B.~Schoelkopf}, \bibinfo{editor}{C.~Burges},
  \bibinfo{editor}{A.~Smola} (Eds.), \bibinfo{booktitle}{Advances in Kernel
  Methods - Support Vector Learning}, \bibinfo{publisher}{MIT Press},
  \bibinfo{year}{1998}.
%Type = Article
\bibitem[{Keerthi et~al.(2001)Keerthi, Shevade, Bhattacharyya, and
  Murthy}]{Keerthi2001}
\bibinfo{author}{S.~Keerthi}, \bibinfo{author}{S.~Shevade},
  \bibinfo{author}{C.~Bhattacharyya}, \bibinfo{author}{K.~Murthy},
\newblock \bibinfo{title}{Improvements to platt's smo algorithm for svm
  classifier design},
\newblock \bibinfo{journal}{Neural Computation} \bibinfo{volume}{13}
  (\bibinfo{year}{2001}) \bibinfo{pages}{637--649}.

\end{thebibliography}

\end{document}